%% file: root.tex
\documentclass[journal]{IEEEtran}
\usepackage{hyperref}

\usepackage{algorithm}
\usepackage{algpseudocode}
\usepackage{amsmath}
\usepackage{amssymb}
\usepackage{booktabs}
\usepackage{multirow}
\usepackage{needspace}
\usepackage{adjustbox}
\usepackage[small]{caption}
\usepackage{bm}
\usepackage{todonotes}
\usepackage{fancyhdr}
\usepackage[linesnumbered,ruled,vlined,algo2e]{algorithm2e}
\newcommand{\myparagraph}[1]{\noindent\textbf{#1}~}
\newcommand{\shortname}{UniPred}

\IEEEoverridecommandlockouts  

\definecolor{jinqired}{HTML}{990000}
\definecolor{todo}{HTML}{FF3B30}

\newcounter{RNum}

\newtheorem{definition}{Definition}
\newcommand{\fref}[1]{Figure~\ref{#1}}
\newcommand{\sref}[1]{Section~\ref{#1}}
\newcommand{\tref}[1]{Table~\ref{#1}}

\ifCLASSINFOpdf

\else

\fi

\begin{document}
%
\title{Unifying Deep Predicate Invention with \\Pre-trained Foundation Models}
%
%

\input{authors}

\markboth{IEEE Transactions on Robotics}{Wang et al.: UniPred}

\maketitle

\begin{abstract}
Long-horizon robotic tasks are hard due to continuous state-action spaces and sparse feedback. Symbolic world models help by decomposing tasks into discrete predicates that capture object properties and relations. Existing methods learn predicates either top-down, by prompting foundation models without data grounding, or bottom-up, from demonstrations without high-level priors. We introduce UniPred, a bilevel learning framework that unifies both. UniPred uses large language models (LLMs) to propose predicate effect distributions that supervise neural predicate learning from low-level data, while learned feedback iteratively refines the LLM hypotheses. Leveraging strong visual foundation model features, UniPred learns robust predicate classifiers in cluttered scenes. We further propose a predicate evaluation method that supports symbolic models beyond STRIPS assumptions. Across five simulated and two real-robot domains, UniPred achieves $2\sim4\times$ higher success rates than top-down methods and $3\sim4\times$ faster learning than bottom-up approaches, advancing scalable and flexible symbolic world modeling for robotics. For more information and project materials (video, code...)   visit \url{https://unipred.github.io/}
\end{abstract}


\IEEEpeerreviewmaketitle

\input{intro}

\input{relatedwork}

\input{problem}

\input{method}
\input{results}

\input{conclusion}

\input{acknowledgment}

\bibliographystyle{IEEEtranBST/IEEEtran}
\bibliography{IEEEtranBST/IEEEabrv,references}

\input{biographies}

\end{document}

%% file: authors.tex

\author{
Qianwei~Wang$^{1,2,*}$,
Bowen~Li$^{1,*}$,
Zhanpeng~Luo$^{1,3}$,
Yifan~Xu$^{2}$,
Vineet~Kamat$^{6}$,
Carol~Menassa$^{6}$,
Alexander~Gray$^{4}$,
Tom~Silver$^{5}$,
Sebastian~Scherer$^{1}$,
Katia~Sycara$^{1}$,
and~Yaqi~Xie$^{1}$
\thanks{$^{*}$Qianwei Wang and Bowen Li contributed equally to this work.
Yaqi Xie is the corresponding author
(e mail: \texttt{yaqixie1996@gmail.com}).}
\thanks{$^{1}$Robotics Institute, Carnegie Mellon University,
Pittsburgh, PA 15213 USA.}
\thanks{$^{2}$Computer Science and Engineering,
University of Michigan, Ann Arbor, MI 48109 USA.}
\thanks{$^{3}$Department of Computer Science,
University of Pittsburgh, Pittsburgh, PA 15260 USA.}
\thanks{$^{4}$Centaur AI Institute,
69 Lincoln Blvd, Ste A240, Lincoln, CA 95648 USA.}
\thanks{$^{5}$Department of Electrical and Computer Engineering,
Princeton University, Princeton, NJ 08544 USA.}
\thanks{$^{6}$Department of Civil and Environmental Engineering,
University of Michigan, Ann Arbor, MI 48109 USA.}
\thanks{Part of this work was conducted while Qianwei Wang and
Zhanpeng Luo were Robotics Institute Summer Scholars affiliated
with the Advanced Agent Robotics Technology Lab at Carnegie Mellon University.}
}

%% file: intro.tex
\section{Introduction}
\label{sec:intro}

\begin{figure}[!t]
	\centering
	\includegraphics[width=\columnwidth]{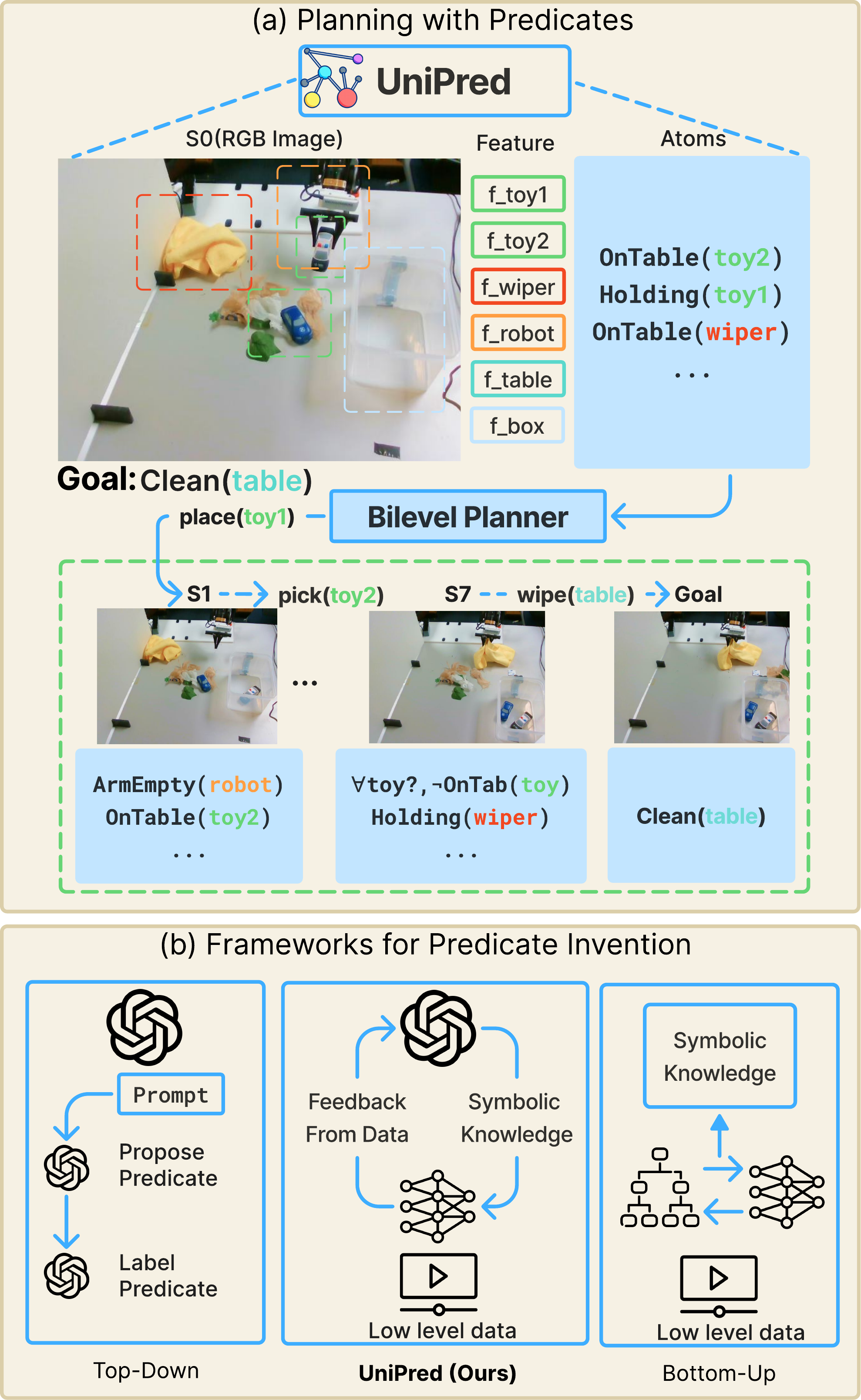}
	\caption{(a) Planning Framework: Given an initial RGB image, UniPred extracts object-centric features using a foundation model (e.g., DinoV3~\cite{simeoni2025dinov3}) and abstracts them into \textbf{invented} ground atoms. 
    These atoms are input into the learned Bilevel Planner to generate a hybrid plan toward the goal. 
    In intermediate steps, UniPred infers corresponding atoms from observed images to facilitate potential replanning.
    (b) Predicate Invention: UniPred effectively invents predicates by unifying top-down knowledge from the foundation model~\cite{achiam2023gpt} with bottom-up feedback derived from data.
    }
	\label{fig:teaser}
\end{figure}

\IEEEPARstart{F}{or} a robot that must make decisions over hours or days, it is neither practical nor necessary to predict every pixel-level detail of world transitions~\cite{mao2024planning,li2024logicity,asai2021latplanpddl}.
Instead, it should plan and reason at a higher level of abstraction.
Consider the example in \fref{fig:teaser}: when learning to clean up a cluttered table, the exact color or type of each toy is irrelevant to high-level decisions.
One thing truly matters is whether the toys have all been packed into the box so that the table is ready to be wiped.
By capturing such essential concepts, symbolic world models~\cite{konidaris2018symbols,chitnis2021nsrt,chitnis2021glib} offer an efficient and generalizable framework for long-horizon tasks, allowing the robot to reason about what needs to be done before deciding how to do it.

Bilevel planning~\cite{chitnis2021nsrt,silver2021operator,silver2022skills,silver2023predicateinvent,liang2024visualpredicator,kumar2023predict,kumar2024practice,li2023embodied,li2025IVNTR} provides a hierarchical framework that exploits the relational structure of symbolic world models for long-horizon, generalizable decision-making.
In this paradigm, low-level continuous states (e.g., RGB images) are abstracted into high-level discrete symbols using predicates (e.g., \texttt{Clean(?table)}).
Planning then proceeds jointly in the symbolic space—deciding what to do—and the continuous space—deciding how to do it.
By leveraging AI planning principles~\cite{helmert2006fast,hoffmann2001ff}, such a bilevel structure achieves greater efficiency and generalization than planning solely in the low-level space.
Nonetheless, most classical planning systems~\cite{kaelbling2011htn,garrett2021integrated} require exhaustive manual engineering of the symbolic model, which largely limits their scalability.
To address this problem, recent work has explored how to learn the abstractions required for bilevel planning—such as predicates~\cite{silver2023predicateinvent,li2023embodied,li2025IVNTR}, PDDL operators~\cite{silver2021operator,kumar2023predict}, samplers~\cite{kumar2024practice}, and skills~\cite{silver2022skills}—directly from demonstrations.

Among these abstractions, \textit{predicates} form the core component.
As shown in \fref{fig:teaser} (a), at the high level, they define the preconditions and effects of planning operators, providing a relational reasoning space in which an AI planner can generate novel solutions to unseen problems~\cite{silver2021operator}.
At the low level, predicates specify the constraints that samplers must satisfy~\cite{garrett2020pddlstream} and determine the initiation and termination conditions of motor skills~\cite{konidaris2018skills}.
Broadly, two paradigms exist for learning symbolic predicates and the associated classifiers.
Top-down approaches~\cite{liang2024visualpredicator,athalye2024predicate,han2024interpret} generate predicates in a zero-shot manner using pre-trained foundation models—such as large language models (LLMs)~\cite{achiam2023gpt,team2023gemini} and visual foundation models (VFMs)~\cite{openai2023gpt4v,simeoni2025dinov3}—leveraging their embedded common-sense priors.
However, because these models are trained on broad, general-purpose knowledge, their zero-shot application to specialized robotic domains typically requires carefully engineered, domain-specific prompts~\cite{llm3,kambhampati2024llms,curtis2024trustproc3ssolvinglonghorizon,valmeekam2022large}. 
Moreover, purely foundation-model-driven methods~\cite{liang2024visualpredicator,athalye2024predicate} often struggle in cluttered scenes because their predicate proposal and classification pipelines rely entirely on test-time prompts, making them unstable to scene variability.
Bottom-up approaches~\cite{silver2023predicateinvent,li2025IVNTR,asai2021latplanpddl,ahmetoglu_2025}, in contrast, directly optimize predicates from low-level data without relying on priors from foundation models.
They are also generally applicable to diverse and noisy states, such as cluttered tabletop scenes in \fref{fig:teaser}, but suffer from low efficiency due to the combinatorial complexity of the predicate search space~\cite{mao2024planning,li2025IVNTR}.

To unify these two families of approaches, we present a highly integrated bilevel learning framework, \shortname, for deep predicate invention.
Since LLMs are often unreliable without carefully engineered prompts~\cite{kambhampati2024llms}, we use them only as coarse top-down guidance, with low-level learning providing bottom-up feedback.
Specifically, \shortname\ first constructs a candidate pool of predicates: the LLM proposes rough predicate–effect distributions, which are used to derive pseudo-labels for training neural predicate classifiers on demonstration data~\cite{li2025IVNTR}.
The classifier’s training loss then serves as feedback that informs the LLM about potential inconsistencies and steers its subsequent proposals toward more plausible hypotheses.
Through this iterative querying process, \shortname\ efficiently generates the predicate candidate set while {reducing reliance on domain-specific prompt engineering}.
Building on top of this bilevel learning framework, \shortname\ further proposes to \textit{optimize} predicate classifiers on the strong feature representations from the recent VFMs~\cite{simeoni2025dinov3} in image-based domains, yielding robust predicate grounding over cluttered real images.
In the second predicate selection stage, prior approaches~\cite{liang2024visualpredicator,athalye2024predicate,li2025IVNTR,silver2023predicateinvent} often fail to distinguish between basic predicates directly grounded by (neural) classifiers (e.g., \texttt{OnTable(?toy)}) and the derived ones (e.g., $\exists$\texttt{?toy,OnTable(?toy)}), leading to unreliable selection in domains with Non-STRIPS operators~\cite{mao2022pdsketch} such as \texttt{WipeTable}.
\shortname\ addresses this by adding only basic predicates in operator effects and treating derived predicates as derived relational constraints when computing planning-driven predicate scores, resulting in a robust symbolic world model in complex, Non-STRIPS domains.

\looseness = -1
We conduct comprehensive evaluation of \shortname\ on 7 robot planning domains with various state representations.
Thanks to the discovered relational symbolic world model, \shortname\  generalizes significantly better than various model-free baselines.
Compared with purely top-down methods~\cite{athalye2024predicate,liang2024visualpredicator}, \shortname\ {reduces the burden of carefully engineered prompt design} and flexibly learns more effective predicates that ground in various state representations.
Compared with purely bottom-up methods~\cite{li2025IVNTR,silver2023predicateinvent}, \shortname\ achieves $3\sim4\times$ speed up with perceptually generalizable predicates on real-world images.
We summarize our key contributions as follows:
\begin{itemize}
    \item \textbf{Foundation Model-based Unified Bilevel Learning:} \shortname\ introduces an iterative framework that queries LLMs with feedback from neural predicate classifiers, thereby {reducing reliance on domain-specific prompt engineering} and substantially improving learning efficiency.
    \item \textbf{Derived-aware Predicate Selection:} 
    \shortname\ differentiates between basic and derived predicates during predicate selection, treating derived predicates as derived relational constraints for predicate evaluation, enabling effective world modeling in Non-STRIPS domains. {This module is separable from the LLM-guided proposal loop and can be applied to predicate candidates generated by other predicate invention methods, such as IVNTR.}
    \item \textbf{Comprehensive Evaluation:} 
We evaluate \shortname\ on 7 domains, including 5 simulated domains with diverse state representations (pose, image, and point cloud) and 2 real-world domains covering 20 tasks in total. {The results provide empirical evidence that \shortname\ exhibits a certain degree of generalization across different task settings while maintaining efficient learning and planning.}

\end{itemize}

%% file: relatedwork.tex
\section{Related Work}
\label{sec:relatedwork}

\subsection{Foundation Models for Robot Planning}

Foundation models support robot planning at multiple levels. At the motion level, vision-language-action (VLA) models trained on extensive robot trajectories map visual and linguistic inputs directly to joint-level actions. They can serve as general-purpose skill primitives alongside classical motion planners for local control or waypoint following~\cite{kim2024openvla,rdt2024rdt,rdt2025rdt2,intelligence2024pi0,intelligence2025pi05}. At the task level, large language models (LLMs) translate goals into subtask sequences, executable code, or symbolic operators, with systems grounding language in robot affordances, learning PDDL-style domains from feedback, or generating control code and task graphs from instructions~\cite{han2024interpret,saycan,code_as_policy,huang2023voxposer,shah2022lmnav}. Visual foundation models, including CLIP and large multimodal transformers, enable open-vocabulary scene understanding and affordance prediction. They supply symbolic state descriptions, semantic cost maps, or object-centric features to downstream planners~\cite{openai2023gpt4v,simeoni2025dinov3,radford2021clip,driess2023palme}. These models thus supply complementary semantic priors, reusable skills, and scene abstractions. Within task and motion planning (TAMP), foundation models have been used to specify subgoals~\cite{yang2024guidinglonghorizontaskmotion}, formulate constraints~\cite{kumar2024openworld}, and ground semantic state abstractions~\cite{athalye2024predicate}. We focus on using their priors and representations to discover predicates for bilevel planning~\cite{athalye2024predicate,silver2023predicateinvent,li2025IVNTR}.

\subsection{Learning for Bilevel Planning}
Classical TAMP methods~\cite{kaelbling2011htn,garrett2020pddlstream,garrett2021integrated} rely on hand-engineered domain models, motivating learning the components required for planning. Bilevel planning addresses TAMP by alternating between a high-level task planner and a low-level sampler. Given a task plan specifying \textit{what} to do, the sampler determines \textit{how} to execute it by choosing continuous parameters satisfying the requisite geometric constraints. Prior work learns planning operators~\cite{chitnis2021nsrt,silver2021operator,kumar2023predict}, samplers~\cite{kumar2024practice}, parametrized skill policies~\cite{silver2022skills}, and relational predicates~\cite{silver2023predicateinvent,liang2024visualpredicator,li2025IVNTR,athalye2024predicate}. Relational representations support generalization to compositionally novel problems, while task decomposition improves long-horizon planning efficiency over flat policies~\cite{zhao2023learning} or planning directly in low-level states. Among these learned abstractions, discovering a rich and expressive set of predicates for low-level states remains particularly challenging~\cite{li2025IVNTR,mao2024planning}. Our framework uses foundation-model priors and representations~\cite{achiam2023gpt,simeoni2025dinov3} to address this challenge. By learning the abstractions used by the planner, it seeks to retain these benefits of structured planning while reducing manual domain engineering.

\subsection{Predicate Discovery for Planning}
Predicates abstract continuous observations, such as poses, point clouds, and images, into high-level discrete states represented by logical atoms (e.g., \texttt{OnTable(?toy)} and \texttt{InBox(?toy)}). These abstractions enable AI planners to search the discrete state space efficiently and generate task plans for novel object compositions~\cite{helmert2006fast} and define the geometric constraints that continuous samplers must satisfy at each planning step.

Top-down methods~\cite{liang2024visualpredicator,athalye2024predicate,han2024interpret} propose predicates using pre-trained foundation models in a zero-shot manner. Their commonsense priors offer data efficiency, but general internet-scale pre-training can be unreliable in specialized domains. These methods are sensitive to prompt design and may propose incomplete predicate sets for unfamiliar tasks~\cite{athalye2024predicate}. Frozen, prompt-based classifiers can also be brittle in cluttered or unseen states~\cite{athalye2024predicate,liang2024visualpredicator}.

Bottom-up methods~\cite{mao2024planning,asai2021latplanpddl,silver2023predicateinvent,li2025IVNTR} instead optimize predicates directly from low-level data. They are typically prompt-free and adaptable to specialized domains and noisy observations, but searching the combinatorial predicate space is costly, particularly for high-dimensional images. Grammar-based methods~\cite{silver2023predicateinvent} are restricted to simple states with explicit relational structures. IVNTR~\cite{li2025IVNTR} learns neural predicate classifiers using an effect-centric objective, offering greater flexibility. However, learning from scratch remains slow in complex domains, and its planning models require all operators to satisfy the STRIPS assumption.

\shortname\ combines foundation-model priors with domain-specific learning feedback to explore predicates efficiently while reducing reliance on domain-specific prompt engineering. Its derived-aware selection procedure distinguishes basic from derived predicates during operator model construction, supporting expressive, planner-friendly abstractions in Non-STRIPS domains. This distinction complements the proposal loop: the loop guides candidate discovery, while selection evaluates predicates for their usefulness in planning. Together, these components support bilevel planning across simulated and real-robot domains.

%% file: problem.tex
\section{Problem Formulation}
\label{sec:problem}
\begin{figure}[!t]
	\includegraphics[width=\columnwidth]{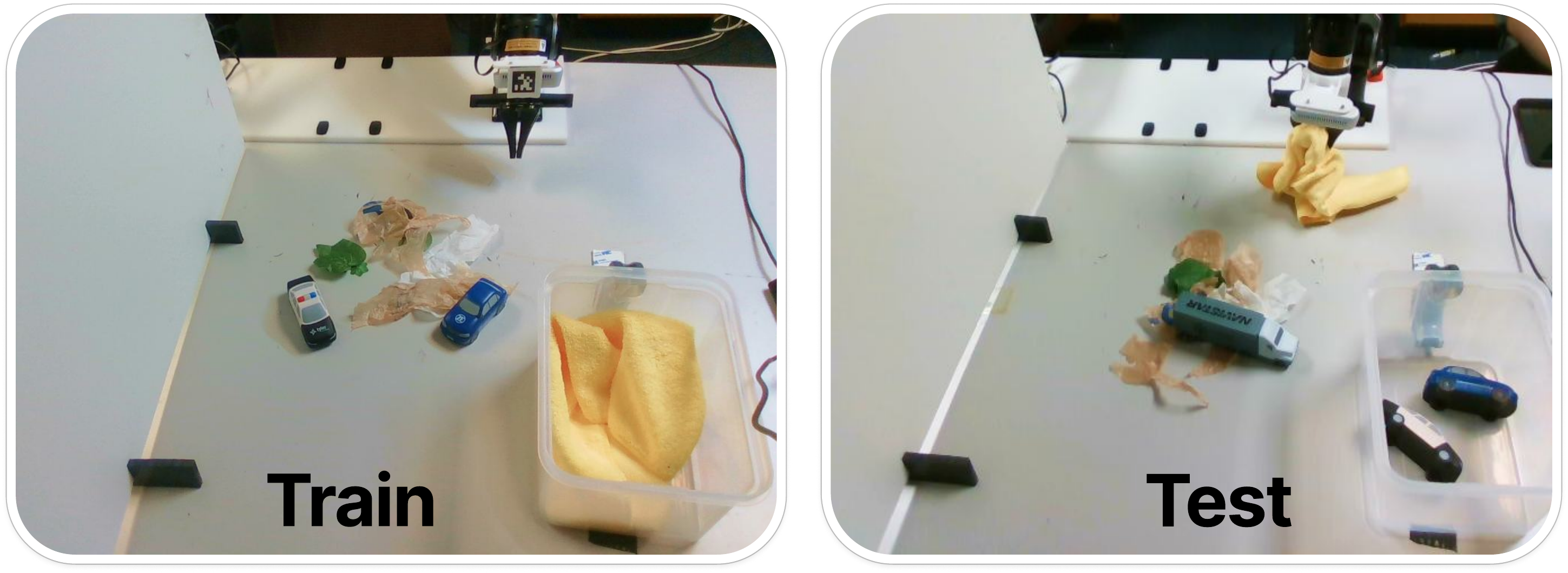}
	\caption{An example of train and test tasks in the table-cleaning domain. During training, both of the two toys are initially on table and the wiper is in the box. During test, the system needs to plan with more toys whose configurations are randomly initialized.}
	\label{fig:domain_example}
\end{figure}

We address the problem of learning planning \emph{abstractions} from offline demonstration datasets.
\shortname\ is designed to (1) generalize to test tasks involving unseen object compositions and (2) maintain computational efficiency during training.
In this section, we formalize the planning domain, action space, predicates, data structure, and computational objective.
Our notation follows~\cite{silver2023predicateinvent,li2025IVNTR} and is summarized in Table~\ref{tab:notation}.

We define a \emph{planning domain} as a tuple
$\mathcal{D} = \langle \Lambda, \mathcal{C}, f, \Psi_\mathrm{G}, \Psi_\mathrm{sta} \rangle$
associated with a task distribution $\mathcal{T}$.
A specific \emph{planning task} is sampled as
$T = \langle \mathcal{O}, \mathbf{x}_0, g \rangle \sim \mathcal{T}$. Here,
$\Lambda$ denotes a finite set of object \emph{types} $\lambda \in \Lambda$, and
$\mathcal{O} = \{\mathtt{o}_1,\dots,\mathtt{o}_N\}$ is a set of $N$ objects, each assigned a type from $\Lambda$.
The task tuple includes the initial continuous state $\mathbf{x}_0$
and a set of ground goal predicates $g \subseteq \Psi_\mathrm{G}$ to be satisfied upon completion.
A state $\mathbf{x} \in \mathcal{X}$ assigns continuous features to every object in $\mathcal{O}$.
For notational simplicity, we represent the state of $N$ objects as a matrix
$\mathbf{x} \in \mathbb{R}^{N \times K}$, where each row encodes $K$ domain-specific features
(e.g., pose, gripper status, or table dimensions), consistent with~\cite{silver2023predicateinvent,li2025IVNTR}.
In our experiments, we extend this formulation to richer object-centric representations, such as point clouds and RGB images, by treating their encodings as per-object features.

The action space comprises a finite set of $M$ \emph{parameterized controllers}, denoted
$\mathcal{C} = \{\mathtt{C}_1,\dots,\mathtt{C}_M\}$.
Each controller $\mathtt{C} \in \mathcal{C}$ is defined by a typed signature
$(\lambda_1,\dots,\lambda_v)$ over object types in $\Lambda$ and a continuous parameter space $\Omega$.
A \emph{grounded} controller instance,
$\underline{\mathtt{C}} = \mathtt{C}(\mathtt{o}_{i_1},\dots,\mathtt{o}_{i_v}, \omega)$, is obtained by binding all object arguments and selecting a parameter $\omega \in \Omega$.
States and actions are linked via a known transition function
$f(\mathbf{x}, \underline{\mathtt{C}}) \mapsto \mathbf{x}'$.
A \emph{plan} for task $T$ is a sequence of grounded controllers
$\pi = [\underline{\mathtt{C}}_1,\dots,\underline{\mathtt{C}}_H]$.
A plan is valid if recursively applying the transition model
$\mathbf{x}_i = f(\mathbf{x}_{i-1}, \underline{\mathtt{C}}_i)$
starting from $\mathbf{x}_0$ yields a terminal state $\mathbf{x}_H$ that satisfies the goal set $g$.

We use a table clean domain as a running example (see Figure~\ref{fig:domain_example}).
This domain includes five object types:
$\Lambda = \{\mathrm{robot}, \mathrm{toy}, \mathrm{box}, \mathrm{wiper}, \mathrm{table}\}$.
The objective is for the robot to pick all $\mathrm{toy}$ objects from the $\mathrm{table}$, place them into a designated $\mathrm{box}$, and subsequently wipe the $\mathrm{table}$ surface.
The controller set for this domain includes primitives such as
\texttt{PickToyFromTable(?r,?to,?ta)},
\texttt{PlaceToyInBox(?r,?t,?b)},\texttt{PushBox(?r,?b)}, and \texttt{WipeTable(?r,?w,?t)}.
These controllers encompass skills derived from motion planning~\cite{silver2023predicateinvent,li2025IVNTR}
(e.g., pick and place) and imitation learning~\cite{zhao2023learning,chi2023diffusionpolicy}
(e.g., \texttt{PushBox}).
We provide further implementation details in the \sref{sec:results}.

To facilitate efficient planning, bilevel methods use a predicate set $\Psi$ to decompose long-horizon decision problems.
A \emph{lifted predicate} is defined by a typed signature over variables,
$\psi(\mathtt{?\lambda}_1,\dots,\mathtt{?\lambda}_u)$,
and an associated classifier
$\theta_\psi : \mathcal{X} \times \mathcal{O}^u \rightarrow \{\mathrm{True}, \mathrm{False}\}$.
A \emph{ground predicate} $\underline{\psi}$ is formed by instantiating all variables with specific objects from $\mathcal{O}$; for brevity, we define
$\theta_{\underline{\psi}}(\mathbf{x}) \triangleq
\theta_\psi(\mathbf{x}, (\mathtt{o}_1,\dots,\mathtt{o}_u))$.
This classifier is used to determine whether the ground predicate holds in state $\mathbf{x}$ for the specific object tuple.

Following~\cite{silver2023predicateinvent,li2025IVNTR}, we categorize predicates into three subsets:
(i) $\Psi_\mathrm{sta}$, static predicates with truth values that remain fixed throughout the task, depending solely on time-invariant object attributes (e.g., \texttt{ColorRed(?toy)} or \texttt{SizeSmall(?toy)};
(ii) $\Psi_\mathrm{dyn}$, dynamic predicates whose truth values may change during execution (e.g., \texttt{InBox(?toy,?box)} or \texttt{OnTable(?toy,?table)}; and
(iii) $\Psi_\mathrm{G}$, goal predicates.
We further partition dynamic predicates into
\emph{basic} dynamic predicates, $\Psi_\mathrm{dyn}^\mathrm{b}$, which are evaluated directly from object-centric features (e.g. \texttt{OnTable(?toy,?table)}),
and \emph{derived} dynamic predicates, $\Psi_\mathrm{dyn}^\mathrm{d}$, which are computed from basic predicates via quantifiers.
For instance, the predicate
$\forall\,\texttt{?toy}.\;\neg \texttt{OnTable(?toy,?table)}$
signifies that no toys remain on the table.
We assume $\Psi_\mathrm{G}$ and $\Psi_\mathrm{sta}$ are provided, whereas the dynamic predicate set $\Psi_\mathrm{dyn}$ may be expanded by the learning algorithm, as in~\cite{li2025IVNTR}.

During training, the robot utilizes an offline demonstration dataset of $B$ task--plan pairs,
$\mathcal{D} = \{(T_b, \pi_b)\}_{b=1}^{B}$, where
$T^b = \langle \mathcal{O}^b, \mathbf{x}_0^{b}, g^b \rangle \sim \mathcal{T}^\mathrm{train}$.
Since $f$ is known and deterministic, we derive the full state--action trajectory
$(\mathbf{x}_0^b, \underline{\mathtt{C}}_1^b, \mathbf{x}_1^b, \dots)$ for each instance by simulating $\pi_b$.
At test time, the agent encounters tasks
$T = \langle \mathcal{O}, \mathbf{x}_0, g \rangle \sim \mathcal{T}^\mathrm{test}$
characterized by unseen object configurations, larger object sets, and more diverse action compositions than those in $\mathcal{T}^\mathrm{train}$.
As illustrated in Figure~\ref{fig:domain_example}, training demonstrations consistently initialize with the robot hand empty, two toys on the table, and the wiper inside the box.
Conversely, test tasks may initialize with the robot holding the wiper and involve three toys in varied locations, such as inside the box.
These variations alter both the requisite plan length and the sequencing of controllers.
Consequently, success necessitates abstractions that capture reusable structure rather than simply memorizing demonstration trajectories.

\begin{table}[!t]
    \centering
    \setlength{\tabcolsep}{2pt}
    \renewcommand{\arraystretch}{1.14}
    \footnotesize
    \caption{{Key notations used throughout this work.}}
    \begin{tabular}{@{}p{0.17\columnwidth}p{0.78\columnwidth}@{}}
        \toprule
        \textbf{Symbol} & \textbf{Meaning} \\
        \midrule
        $\mathcal{D}$ & Planning domain or offline demonstration dataset, clear from context \\
        $\mathcal{T}$, $T$ & Task distribution and sampled planning task \\
        $\mathcal{O}$, $\mathbf{x}$ & Object set and object-centric continuous state \\
        $\mathcal{C}$, $\mathtt{C}$ & Controller set and parameterized controller \\
        $\underline{\mathtt{C}}$ & Grounded controller instance \\
        $f$ & State transition function \\
        $\pi$, $\bar{\pi}$ & Refined plan and symbolic plan skeleton \\
        $\psi$, $\underline{\psi}$ & Lifted predicate and its grounded instance \\
        $\theta_\psi$ & Neural classifier for predicate $\psi$ \\
        $\Psi_\mathrm{sta}$, $\Psi_\mathrm{G}$ & Static predicates and goal predicates \\
        $\Psi_\mathrm{dyn}^\mathrm{b}$, $\Psi_\mathrm{dyn}^\mathrm{d}$ & Basic and derived dynamic predicates \\
        $\mathtt{Op}^{\mathtt{C}}$, $\eta^{\mathtt{C}}$ & Operator and sampler for a controller \\
        $\hat{\Psi}_\mathrm{dyn}$ & Candidate dynamic predicate set \\
        $\Delta^\psi$, $\bm{t}^{\psi,\underline{\mathtt{C}}}$ & Lifted and grounded effect vectors for predicate $\psi$ \\
        $\tau$ & Total learning time \\
        \bottomrule
    \end{tabular}
    \label{tab:notation}
\end{table}

%% file: method.tex
\section{Methodology}

\label{sec:methodology}
\begin{figure*}[!t]
    \centering
    \includegraphics[width=2\columnwidth]{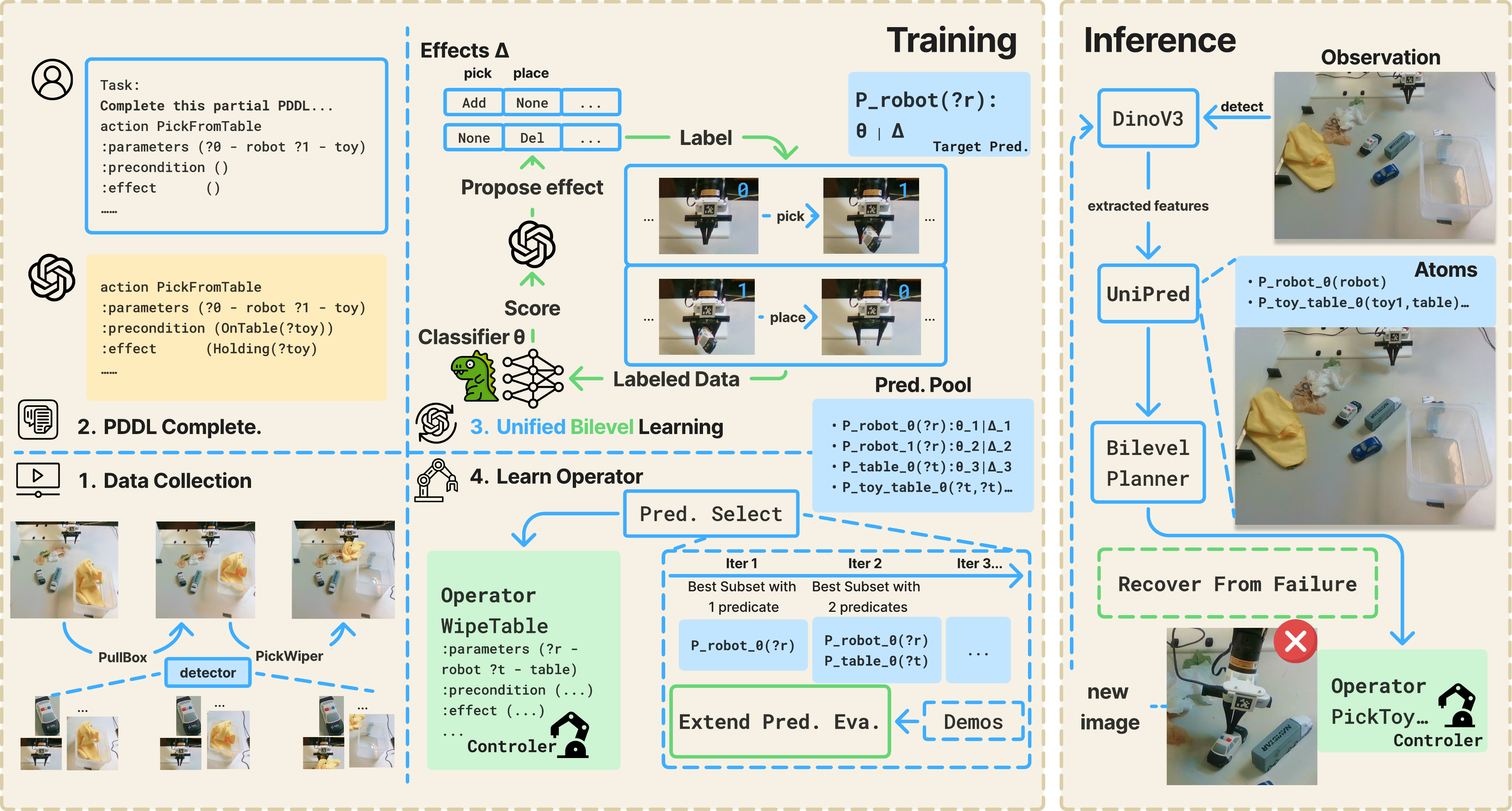}
    \caption{
    Overview of the \shortname\ Framework: Training and Inference. 
    The framework consists of two main phases. 
    Training (Left) utilizes an offline demonstration dataset, where we first prompt a Large Language Model (LLM) to complete a partial PDDL domain specification. 
    This is followed by an LLM-in-the-loop bilevel learning framework designed to construct a comprehensive set of predicate candidates, which are then refined through a final down-selection step. 
    During Inference (Right), \shortname\ processes input images by extracting robust visual features from object-centric regions using a VFM~\cite{simeoni2025dinov3}, which are converted to the initial set of ground predicates. 
    Based on the predicates, the learned bilevel planner then generates the initial execution plan and optionally replans when skill execution failures are observed.
    }
    \label{fig:pipeline}
\end{figure*}

\subsection{Bilevel Planning with Predicates}
We provide a brief review of bilevel planning and refer the reader to existing literature for a comprehensive discussion~\cite{silver2021operator,silver2022skills,silver2023predicateinvent,kumar2023predict,kumar2024practice,li2023embodied,li2025IVNTR,garrett2021integrated}.
Figure~\ref{fig:teaser}a illustrates an instance of our setting within the table-cleaning domain.

Bilevel planning leverages relational abstractions to achieve sequential and compositional generalization.
The framework relies on two primary abstractions: \emph{predicates} (state abstractions) and \emph{operators} (action abstractions).
Furthermore, bilevel planning employs relational \emph{samplers} to generate continuous parameters for controllers.

\begin{definition}[Operator]\label{def:op}
The \textit{operator} for a parameterized controller $\mathtt{C}$ is a tuple $\mathtt{Op}^\mathtt{C} = \langle \mathtt{Var}, \mathtt{Pre}, \mathtt{Eff}^+, \mathtt{Eff}^- \rangle$, where $\mathtt{Var}$ is a tuple of object placeholders matching the type signature of $\mathtt{C}$, and $\mathtt{Pre},\,\mathtt{Eff}^+,\,\mathtt{Eff}^- \subseteq \Psi$ are sets of lifted predicates defined over variables in $\mathtt{Var}$, representing \emph{preconditions}, \emph{add effects}, and \emph{delete effects}, respectively.
Here, $\Psi$ denotes the predicate set.
\end{definition}
As an illustrative example, the operator for $\mathtt{PickToyFromTable(?r,?t)}$ may be defined as:
\begin{align*}
& \mathtt{Pre}=\{\mathtt{HandEmpty(?r)},\mathtt{ToyOnTable(?t,?t)}\},\\
& \mathtt{Eff}^+=\{\mathtt{Holding(?r,?t)}\},\\
& \mathtt{Eff}^-=\{\mathtt{HandEmpty(?r)},\mathtt{ToyOnTable(?t,?t)}\}.
\end{align*}

Given a task $T = \langle \mathcal{O}, \mathbf{x}_0, g\rangle$, the bilevel planning process (see Figure~\ref{fig:teaser}a) initiates by constructing an \emph{abstract state} comprising all ground predicates over objects $\mathcal{O}$ satisfied in $\mathbf{x}_0$.
These initial ground predicates, alongside the operator set and goal $g$, are input to a classical AI planner~\cite{helmert2006fast} to synthesize a plan \textit{skeleton} $\bar{\pi}$.
This skeleton consists of a sequence of partially grounded actions $\underline{\bar{\mathtt{C}}}$ with unspecified continuous parameters.
To refine the actions in this skeleton $\underline{\bar{\mathtt{C}}}\in\bar{\pi}$ into fully grounded actions $\underline{\mathtt{C}}$ equipped with continuous parameters $\omega$, bilevel planning utilizes \textit{samplers}.

\begin{definition}[Sampler]
The \textit{sampler} $\eta^\mathtt{C}$ for a planning operator $\mathtt{Op}^\mathtt{C}$ with $v$ object placeholders is a conditional distribution
$\omega \sim \eta^\mathtt{C}(\cdot \mid \mathbf{x}, (\mathtt{o}_1, \ldots, \mathtt{o}_{v}))$
that proposes continuous action parameters for $\mathtt{C}((\mathtt{o}_1, \ldots, \mathtt{o}_{v}), \cdot)$ given a state $\mathbf{x}$.
\end{definition}

In contrast to deterministic operators, samplers are typically stochastic and may fail to yield valid parameters.
Consequently, bilevel planning alternates between the AI planner and the samplers, using the predicate classifiers $\theta_\Psi$ to guide the search and compensate for potential sampling failures.

Assuming a sufficiently expressive predicate set, prior work has demonstrated the learnability of \textit{operators}~\cite{chitnis2021nsrt} and \textit{samplers}~\cite{silver2022skills,kumar2024practice} from demonstration data $\mathcal{D}$.
In this context, predicates fulfill two critical functions.
First, the \textit{relational} nature of predicates, learned operators, and samplers enables strong generalization to held-out test tasks ($\mathcal{T}^\mathrm{test}$) involving varying object sets and configurations.
Second, informative \emph{dynamic} predicates structure the abstract state space, allowing the planner to efficiently search the space of operator sequences, as depicted in Figure~\ref{fig:teaser}a.
Conversely, an impoverished predicate set—containing, for instance, only $\Psi_\mathrm{G}$ and $\Psi_\mathrm{sta}$—can render bilevel planning computationally intractable~\cite{silver2023predicateinvent}.
We next introduce the \shortname\ framework, which addresses this challenge by automatically inventing dynamic predicates to facilitate efficient bilevel planning.

\subsection{UniPred Overview}

Figure~\ref{fig:pipeline} summarizes training and inference.

\paragraph{Stage one: Data collection and preparation}
We extract labeled state-transition pairs $(\mathbf{x}, \underline{\mathtt{C}}, \mathbf{x}')$ from demonstrations. In image domains, an object detector~\cite{lang_sam} produces crops that a visual encoder maps to object-centric states.

\paragraph{Stage two and three: Learning a pool of basic dynamic predicates}
An LLM completes a partial PDDL domain to propose seed predicates and effects. Effect-based supervision~\cite{li2025IVNTR} trains their neural classifiers, whose learning feedback guides subsequent proposals. In image domains, classifiers operate on VFM features~\cite{simeoni2025dinov3}; approximately 20 demonstrations suffice in our table-cleaning experiments. Section~\ref{sec:predicate learning} details this loop.

\paragraph{Stage four: Derived-aware predicate selection and operator learning}
A hill-climbing search selects predicates using a planning-driven objective~\cite{silver2023predicateinvent,li2025IVNTR,liang2024visualpredicator}. It distinguishes basic predicates $\Psi_\mathrm{dyn}^\mathrm{b}$ from derived predicates $\Psi_\mathrm{dyn}^\mathrm{d}$, which need not satisfy STRIPS effect assumptions (Section~\ref{sec:predicate selection}). The selected set supports operator, sampler, and skill learning~\cite{chitnis2021nsrt,silver2022skills,kumar2024practice} (Section~\ref{sec:operator_sampler_skill}).

\paragraph{Inference and closed-loop planning}
For an RGB observation, detection and DINOv3 encoding produce one feature vector per object. Learned classifiers ground these features into atoms; a classical planner combines them with the operators and goal predicates $\Psi_\mathrm{G}$ to generate a skeleton $\bar{\pi}$. A sampler or closed-loop skill grounds and executes its first action. After each execution, fresh observations update the atoms to check the next action's applicability. Execution failures trigger replanning, as in the failed grasp in Figure~\ref{fig:pipeline}. This cycle ends when the goal is satisfied or the planning budget is exhausted.

\begin{figure}[!t]
	\includegraphics[width=\columnwidth]{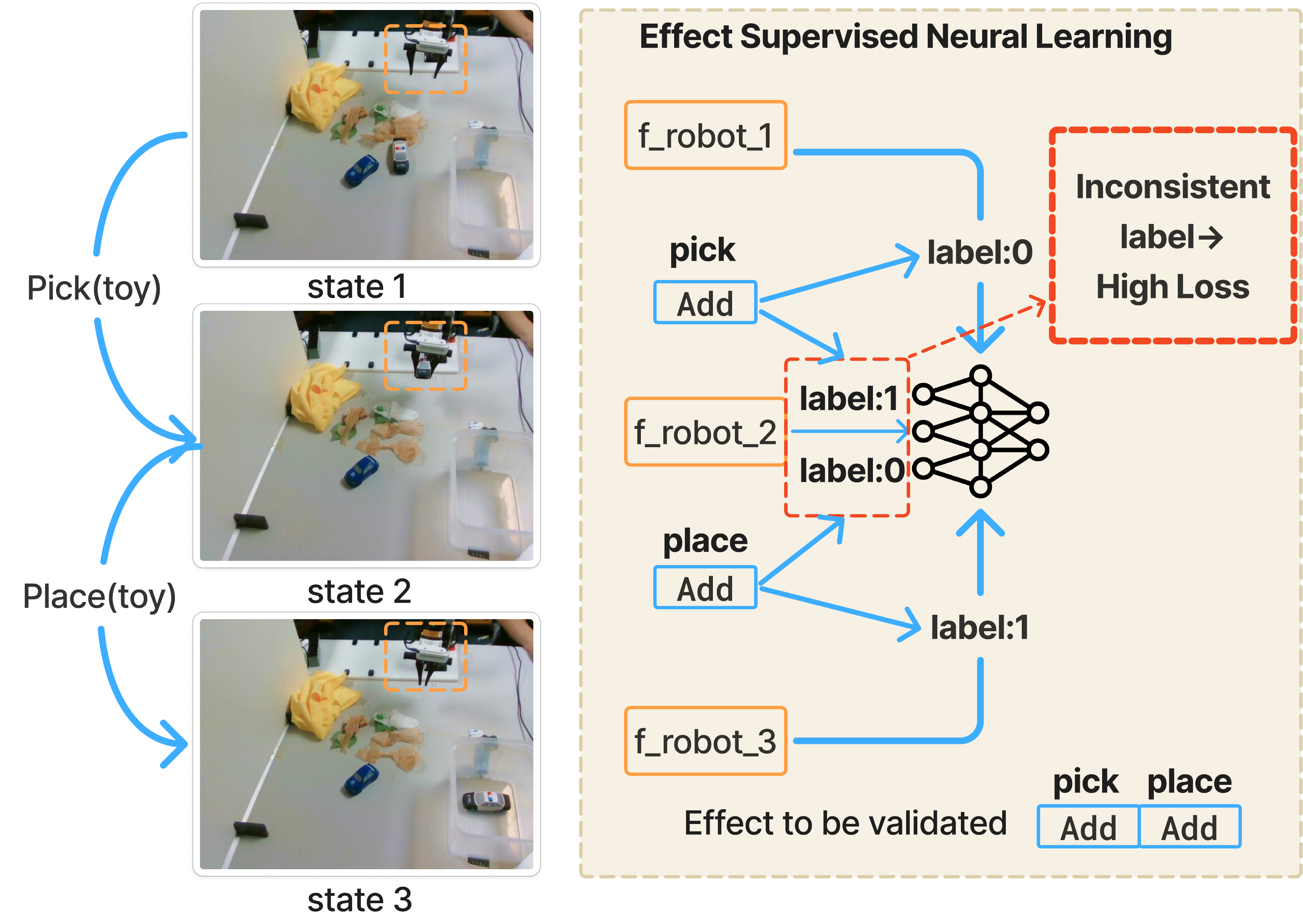}
	\caption{
Effect supervised training process for the classifier of predicate $p_{\text{robot}}(\text{?robot})$. 
The left panel shows three consecutive states during a pick and place sequence, and the right panel illustrates how an incorrect effect specification leads to inconsistent labels across these states, which increases the training and validation loss.
}
	\label{fig:ivntr}
\end{figure}

\subsection{Foundation Model-based Predicate Candidate Learning}
\label{sec:predicate learning}

\begin{figure}[!t]
	\includegraphics[width=\columnwidth]{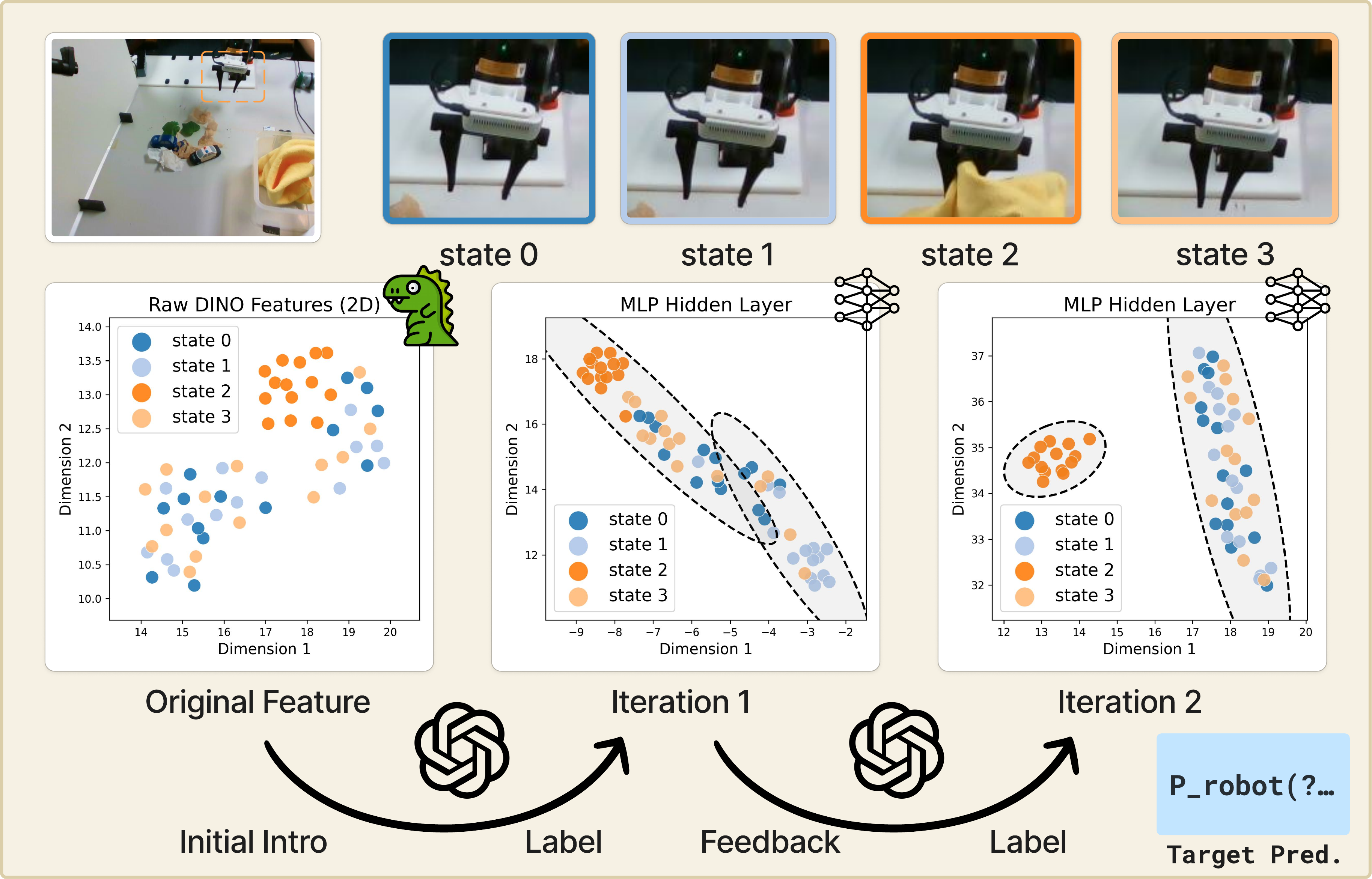}
	\caption{
Overview of unified bilevel learning.
The top row depicts four image states from the table-cleaning task: the robot hand is empty in states 0, 1, and 3, while it holds a towel in state 2.
The bottom row visualizes the evolution of DINO image features through the MLP classifier.
Left: The original 2D projection, where the four states are not clearly separable.
Middle (Iteration 1): After the LLM proposes symbolic knowledge and generates training labels, the hidden features begin to exhibit structure, though state 2 remains partially entangled.
Right (Iteration 2): The task loss is fed back to the LLM, prompting an update to the predicate specification and labels; this yields a refined hidden representation where state 2 is cleanly separated from the empty-hand states.
}
	\label{fig:fm_bilevel}
\end{figure}

Without ground-truth atom labels, \shortname\ trains predicate classifiers using effect-based supervision~\cite{li2025IVNTR} and learns the effects through LLM proposals and low-level feedback.

\paragraph{Effect-based Supervision on Predicate Classifiers}
Let us assume the effect distribution of a predicate $\psi$ within operator effects is known—specifically, whether $\psi$ is added or deleted by each controller class $\mathtt{C}_i \in \mathcal{C}$.
We encode this information in a lifted effect vector $\Delta^\psi = [\delta_1^\psi,\dots,\delta_M^\psi] \in \{-1,0,1\}^M$, where each entry indicates whether $\psi$ constitutes an add effect, a delete effect, or remains unaffected by controller $\mathtt{C}_i$\footnote{Following~\cite{li2025IVNTR}, we assume: (1) each predicate appears at most once in an operator's effect set, and (2) ground operators modify only ground predicates sharing the same or a subset of parameters.}.
Given a transition pair $(\mathbf{x}, \underline{\mathtt{C}}_i, \mathbf{x}')$, the lifted effect vector is instantiated as a ground effect vector $\bm{t}^{\psi,\underline{\mathtt{C}}_i}$. This vector specifies for each ground atom $\underline{\psi}$ over the task objects $\mathcal{O}$ whether its value should transition or remain constant following the application of $\underline{\mathtt{C}}_i$.

As illustrated in Figure~\ref{fig:ivntr}, this effect information provides critical supervision for training the classifier $\theta_\psi$.
For a lifted predicate $\psi$ absent from the effect set of an operator $\mathtt{Op^C}_i$, the truth values of all its ground atoms must remain invariant across transitions induced by any ground controller $\mathtt{C}_i$.
Conversely, if $\psi$ appears in the effect set of $\mathtt{Op^C}_i$, the pre- and post-transition truth values of its ground atoms must reflect the add or delete effects specified by $\Delta^\psi$.
Aggregating these constraints over all transitions and controllers $\mathtt{C} \in \mathcal{C}$ yields a differentiable loss $\mathcal{L}^{\mathcal{D}}(\theta_\psi)$.
This loss is minimized via standard deep learning optimizers to train the neural classifier $\theta_\psi$ directly from object-centric features, eliminating the need for ground truth labels.

This bottom-up, effect-based learning scheme offers a significant advantage over purely top-down approaches~\cite{liang2024visualpredicator,athalye2024predicate} that rely on zero-shot, prompting-based foundation models for predicate classification.
Since foundation models are pre-trained on broad, internet-scale data, their zero-shot classification performance often degrades in specialized or custom domains.
In contrast, the neural predicates $\theta_\psi$ in~\cite{li2025IVNTR} and \shortname\ are optimized directly against low-level state transitions.
This results in predicates that are better aligned with complex environmental states, such as those encountered in cluttered table-cleaning scenarios.

However, identifying the correct effects $\Delta^\psi$ in prior work~\cite{li2025IVNTR} requires a tree expansion algorithm guided by neural learning feedback.
The search space for this method expands exponentially with the number of controllers and objects~\cite{mao2024planning}.
To overcome this efficiency bottleneck, we introduce the LLM-in-the-loop bilevel learning algorithm employed by \shortname.

\paragraph{Unified FM-Bilevel Learning Framework}
Drawing on the complementary strengths of top-down~\cite{liang2024visualpredicator,athalye2024predicate} and bottom-up approaches~\cite{li2025IVNTR}, as well as recent advances utilizing foundation models as optimizers to refine their own outputs~\cite{yang2024large,heni2025sasprompt}, we propose a unified predicate learning framework shown in \fref{fig:pipeline} and \fref{fig:fm_bilevel}.
In \shortname, foundation models serve three distinct roles: (1) initializing the effect distribution by completing the partial domain PDDL; (2) operating in-the-loop with effect-based supervision to generate a consistent pool of dynamic predicates; and (3) In image-based domains, providing image-based feature encodings for the learnable predicate classifier $\theta_\psi$.

First, we represent the set of controllers $\mathcal{C}$ and the limited set of given predicates $\{\Psi_\mathrm{sta}, \Psi_\mathrm{G}\}$ as a structured PDDL~\cite{McDermott1998PDDL} domain definition.
We use this partial domain, supplemented by a small number of demonstrations from $\mathcal{D}$, to prompt an LLM for completion.
From the completed PDDL, we extract newly proposed predicates and their symbolic effect distributions.
For each candidate, we instantiate a neural classifier $\theta_\psi$, train it using the aforementioned effect-based supervision, and record the resulting effect distribution and learning feedback in the exploration history.
These ``seed" predicates encode a prior over the domain structure, enabling the LLM to explore potentially consistent effects more efficiently than the 
\emph{ab initio} search required in~\cite{li2025IVNTR}.

Second, given that initial predicate proposals are typically incomplete, \shortname\ employs the LLM as a proposer and the classifier learning process as an evaluator.
For each initially proposed predicate $\psi$, we query the language model to suggest possible effect patterns based on the current learning history.
\shortname\ instantiates each proposal as a neural classifier $\theta_\psi$, trained via the effect-based loss $\mathcal{L}^{\mathcal{D}}(\theta_\psi)$.
The resulting effect vector and learning feedback are then appended to the history to inform the subsequent iteration of LLM proposals.

Formally, for a single transition pair $(\mathbf{x}, \underline{\mathtt{C}}, \mathbf{x}')$, we apply $\theta_\psi$ to all possible groundings of $\psi$ over the object set $\mathcal{O}$ to obtain:
\[
\hat{\mathbf{v}}, \hat{\mathbf{v}}' = \mathrm{Ground}(\mathbf{x}, \theta_\psi), \ \mathrm{Ground}(\mathbf{x}', \theta_\psi) \in [0,1]^P,
\]
where $P$ denotes the number of ground atoms of $\psi$.
Let $\bm{t}^{\psi,\underline{\mathtt{C}}} \in \{-1,0,1\}^P$ represent the ground effect vector derived from the lifted effect vector $\Delta^\psi$ for controller class $\mathtt{C}$.
We define masks to identify unaffected and affected ground atoms,
\[
\bm{m}^0 = \mathbb{I}\big(\bm{t}^{\psi,\underline{\mathtt{C}}} = 0\big), \quad
\bm{m}^1 = \mathbb{I}\big(|\bm{t}^{\psi,\underline{\mathtt{C}}}| = 1\big),
\]
and let $S^1 = \{p : m^1_p = 1\}$ be the set of indices corresponding to affected atoms.
The per-transition loss is defined as:
\[
\mathcal{L}(\mathbf{x}, \mathbf{x}', \theta_\psi)
=
\mathcal{L}_\mathrm{zero}
+
\mathcal{L}_\mathrm{one},
\]
where
\[
\mathcal{L}_\mathrm{zero}
=
\mathrm{JS}\!\Big(
\hat{\mathbf{v}} \odot \bm{m}^0
\ \Big\|\ 
\hat{\mathbf{v}}' \odot \bm{m}^0
\Big)
\]
penalizes deviations in atoms that should remain constant, and
\[
\mathcal{L}_\mathrm{one}
=
\frac{1}{|S^1|}
\sum_{p \in S^1}
\mathrm{BCE}\Big(
[\hat{v}_p, \hat{v}'_p],
\big[\tfrac{1 - t_p}{2}, \tfrac{1 + t_p}{2}\big]
\Big)
\]
penalizes violations of the add ($t_p=1$) or delete ($t_p=-1$) effects on affected atoms.
Here, $\mathrm{JS}(\cdot\|\cdot)$ denotes the Jensen-Shannon divergence and $\mathrm{BCE}(\cdot,\cdot)$ denotes binary cross-entropy.
The global loss over the dataset is given by:
\[
\mathcal{L}^{\mathcal{D}}(\theta_\psi)
=
\sum_{\mathtt{C} \in \mathcal{C}}
\mathbb{E}_{(\mathbf{x},\underline{\mathtt{C}},\mathbf{x}') \sim \mathcal{D}_\mathtt{C}}
\mathcal{L}(\mathbf{x}, \mathbf{x}', \theta_\psi),
\]
where $\mathcal{D}_\mathtt{C}$ represents the distribution of transitions for the controller class $\mathtt{C}$ in the domain.

After training, we evaluate $\theta_\psi$ on held-out data to obtain $\ell_{\mathrm{val}}(\psi)$. Monotonic rescaling, such as linear normalization across candidates, maps this loss to a score $s(\psi) \in [0,100]$ for LLM feedback. All seed predicates undergo this loop. Proposals below the validation-loss threshold enter the basic predicate pool $\hat{\Psi}_\mathrm{dyn}^\mathrm{b}$; the loop stops at a target number of accepted predicates or a maximum iteration count, both hyperparameters.

In image domains, \shortname\ learns classifiers atop pre-trained DINOv3 representations, unlike fixed grammar functions~\cite{silver2023predicateinvent}, frozen VFM classifiers~\cite{athalye2024predicate}, or classifiers trained from scratch~\cite{li2025IVNTR}. An object detector~\cite{ravi2024sam} produces crops, and the visual encoder maps each to a feature vector, forming $\mathbf{x} \in \mathbb{R}^{N \times K}$. These geometric and semantic features support generalization from sparse demonstrations.

\subsection{Derived-aware Predicate Selection}
\label{sec:predicate selection}
{This selection module takes as input a pool of basic dynamic predicates, which may be generated by \shortname\ or by another predicate invention method such as IVNTR.}
\begin{algorithm2e}[!t]
    \caption{Derived-aware Predicate Selection}
    \label{alg:predicate_selection}
    \DontPrintSemicolon
    \KwIn{Basic predicate pool $\hat{\Psi}_\mathrm{dyn}^\mathrm{b}$, Training tasks $\mathcal{D}$, Goal/Static predicates $\Psi_\mathrm{G}, \Psi_\mathrm{sta}$}
    \KwOut{Selected dynamic predicates $\Psi_\mathrm{dyn}$}
    \BlankLine

    Initialize selected set $\Psi_\mathrm{dyn} \leftarrow \emptyset$\;
    Initialize candidate pool $\tilde{\Psi}_{\mathtt{dyn}} \leftarrow \hat{\Psi}_\mathrm{dyn}^\mathrm{b}$ \;
    Initialize best score $J^* \leftarrow \infty$\;
    \BlankLine

    \While{True}{
        $\psi_{best} \leftarrow \text{None}, \quad J_{min} \leftarrow \infty$\;
        \BlankLine
        
        \For{each candidate $\psi \in \tilde{\Psi}_{\mathtt{dyn}}$}{
            Form candidate set $\tilde{\Psi} \leftarrow \Psi_\mathrm{dyn} \cup \{\psi\} \cup \Psi_\mathrm{G} \cup \Psi_\mathrm{sta}$\;
            \BlankLine
            
            \textbf{Procedure} \textsc{Evaluate}($\tilde{\Psi}, \mathcal{D}$):\;
            \Indp
                Ground states in $\mathcal{D}$ to obtain atom dataset\;
                Learn operators $\tilde{\mathtt{Op}}$ using $\tilde{\Psi}$\;
                \BlankLine
                
                Initialize A* planner with $\tilde{\mathtt{Op}}$\;
                \While{planning for each task in $\mathcal{D}$}{
                    Expand current state using operators\;
                    \BlankLine
                    
                    \If{derived predicates in $\tilde{\Psi}$}{
                        Compute derived atoms
                    }
                    \BlankLine
        
        Reconstruct plan skeleton $\pi$\;
    }
    Compute score $J$ comparing $\pi$ with demonstrations $\bar{\pi}$\;
            \Indm
            \BlankLine
            
            \If{$J < J_{min}$}{
                $J_{min} \leftarrow J, \quad \psi_{best} \leftarrow \psi$\;
            }
        }
        \BlankLine

        \If{$J_{min} \ge J^*$ \textbf{or} $J_{min} < \epsilon$}{
            \textbf{break} {\small\textcolor{lightgray}{\tcp*[r]{Converged or threshold met}}}
        }
        \BlankLine

        $\Psi_\mathrm{dyn} \leftarrow \Psi_\mathrm{dyn} \cup \{\psi_{best}\}$\;
        $\tilde{\Psi}_{\mathtt{dyn}} \leftarrow \tilde{\Psi}_{\mathtt{dyn}} \setminus \{\psi_{best}\}$\;
        $J^* \leftarrow J_{min}$\;
        \BlankLine

        \If{$\psi_{best} \in \hat{\Psi}_\mathrm{dyn}^\mathrm{b}$}{
            Generate derived forms ${\Psi}_{\mathrm{dyn},best}^\mathrm{d}$ from $\psi_{best}$\;
            $\tilde{\Psi}_{\mathtt{dyn}} \leftarrow \tilde{\Psi}_{\mathtt{dyn}} \cup {\Psi}_{\mathrm{dyn},best}^\mathrm{d}$ {\small\textcolor{lightgray}{\tcp*[r]{Expand search space}}}
        }
    }
    \Return $\Psi_\mathrm{dyn}$\;
\end{algorithm2e}

Given the basic dynamic predicate pool $\hat{\Psi}_\mathrm{dyn}^\mathrm{b}$, \shortname\ identifies the subset of predicates most valuable for planning.
Following previous work~\cite{silver2023predicateinvent,li2025IVNTR}, we consider augmenting the basic predicates with negations and quantifiers, which are denoted as derived dynamic predicate candidates $\hat{\Psi}_\mathrm{dyn}^\mathrm{d}$.
Crucially, \shortname\ distinguishes between $\hat{\Psi}_\mathrm{dyn}^\mathrm{b}$ and $\hat{\Psi}_\mathrm{dyn}^\mathrm{d}$, yielding symbolic models that are more expressive than those produced by prior predicate invention approaches.

At a high level, previous methods select the final predicate set $\Psi_\mathrm{dyn}$ from the union of basic and derived predicates $\hat{\Psi}_\mathrm{dyn}^\mathrm{b} \cup \hat{\Psi}_\mathrm{dyn}^\mathrm{d}$ by optimizing a planning-driven surrogate objective $J(\Psi_\mathrm{dyn}, \mathcal{D})$~\cite{silver2023predicateinvent}.
Specifically, $J(\Psi_\mathrm{dyn}, \mathcal{D})$ evaluates whether the planning operators~\cite{chitnis2021nsrt} derived from $\Psi_\mathrm{dyn}$ can: (1) reproduce the demonstrated task plans in $\mathcal{D}$, and (2) minimize node expansions during task planning.
However, a key limitation of prior work~\cite{silver2023predicateinvent,li2025IVNTR} is that this evaluation relies on STRIPS assumptions, where predicates must appear consistently in the effects of operators.
A representative failure case is illustrated in our table cleaning domain (see \fref{fig:teaser}), where the derived predicate $\forall\,\texttt{?toy}.\;\neg P_{\text{toy\_table\_0}}\texttt{(?toy,?table)}$ is a critical precondition for the operator $\mathtt{WipeTable(\mathrm{?table})}$, yet it does not appear as a consistent effect in any operator.
Consequently, the operator learning stage fails during the calculation of the planning-driven surrogate objective, resulting in an unexpressive set of predicates.

A key insight in \shortname\ is that basic dynamic predicates $\hat{\Psi}_\mathrm{dyn}^\mathrm{b}$ naturally appear consistently in operator effects, whereas derived predicates $\hat{\Psi}_\mathrm{dyn}^\mathrm{d}$ may not.
This discrepancy arises because effect-based supervision~\cite{li2025IVNTR} implicitly ensures that for any predicate absent from an operator's effect set, its ground atoms remain constant during transitions.
Since derived candidates $\hat{\Psi}_\mathrm{dyn}^\mathrm{d}$ are not learned with such supervision, they may exhibit inconsistent appearances in effect sets.
Motivated by this observation, we introduce a derived-aware predicate selection pipeline in \shortname.
The complete predicate selection procedure is detailed in Algorithm~\ref{alg:predicate_selection}.

Algorithm~\ref{alg:predicate_selection} starts with an empty selected set and the basic candidate pool. Each hill-climbing iteration evaluates single-predicate additions using $J$, and adds derived forms to the pool whenever a basic predicate is selected. Search stops when the target score is reached or no improvement remains.

For each candidate set $\tilde{\Psi}$, evaluation grounds the demonstration states, constructs operator effects from learned lifted effect vectors $\Delta^\psi$, and learns preconditions by intersection~\cite{chitnis2021nsrt}. An A* planner~\cite{hart1968astar} generates skeletons for the training tasks; comparison with demonstration skeletons $\bar{\pi}$ gives $J(\tilde{\Psi}_\mathrm{dyn}, \mathcal{D})$.

Our key strategy during A* search is to use operators to transition basic ground atoms, and subsequently calculate derived ground atoms from the basic state, rather than predicting them via operators.
At each symbolic state, we first apply an operator to obtain the next set of ground atoms for the basic predicates $\Psi_\mathrm{dyn}^\mathrm{b}$.
We then perform a \textit{derived closure} step that deterministically computes all derived atoms implied by the current basic atoms and the definitions in $\Psi_\mathrm{dyn}^\mathrm{d}$.
Consider the example from the table cleaning domain with the derived predicate $\forall\,\texttt{?toy}.\;\neg P_{\text{toy\_table\_0}}\texttt{(?toy,?table)}$.
Suppose that applying an operator deletes all basic atoms of $P_{\text{toy\_table\_0}}\texttt{(?toy,?table)}$.
During derived closure, the system evaluates the universal condition and adds the corresponding ground atom $\forall\,\texttt{?toy}.\;\neg P_{\text{toy\_table\_0}}\texttt{(?toy,?table)}$ to the state.
Thus, even though the quantifier never appears directly in any operator effect set, this closure step allows the planner to evaluate its truth value based on the basic atoms $P_{\text{toy\_table\_0}}$ and plan accordingly.

Upon termination of the hill-climbing search, the resulting dynamic predicate set $\Psi_\mathrm{dyn} = \Psi_\mathrm{dyn}^\mathrm{b} \cup \Psi_\mathrm{dyn}^\mathrm{d}$ is passed to the operator, sampler, and skills learning stages.

\subsection{Operator, Sampler, and Skill Learning}
\label{sec:operator_sampler_skill}

We primarily build upon prior work on operator and sampler learning for bilevel planning~\cite{li2025IVNTR,silver2022skills,silver2023predicateinvent,kumar2024practice}, extending it to our setting which integrates both parametrized controllers and learned closed-loop skills.

Given the final predicate set $\Psi = \{\Psi_\mathrm{G}, \Psi_\mathrm{sta}, \Psi_\mathrm{dyn}\}$ and the controller set $\mathcal{C}$, the predicate learning phase yields the corresponding lifted effect vectors $\Delta^\psi$ for all $\psi \in \Psi$.
For each controller class $\mathtt{C} \in \mathcal{C}$, we construct its operator
$\mathtt{Op}^\mathtt{C} = \langle \mathtt{Var}, \mathtt{Pre}, \mathtt{Eff}^+, \mathtt{Eff}^- \rangle$
by extracting $\mathtt{Eff}^+$ and $\mathtt{Eff}^-$ from the components of $\Delta^\psi$, and deriving $\mathtt{Pre}$ using the standard intersection rule over ground atoms that hold prior to successful executions of $\mathtt{C}$~\cite{chitnis2021nsrt,silver2023predicateinvent}.

For parametrized controllers, we learn samplers $\eta^\mathtt{C}$ from demonstration tuples $(\mathbf{x}, \mathtt{C}, (\mathtt{o}_1,\dots,\mathtt{o}_v), \omega)$ via supervised learning.
This process models the conditional distribution
$\omega \sim \eta^\mathtt{C}(\cdot \mid \mathbf{x}, (\mathtt{o}_1,\dots,\mathtt{o}_v))$, following the formulation in~\cite{silver2022skills}.

Furthermore, our controller set $\mathcal{C}$ incorporates closed-loop skills learned directly from the same demonstrations (e.g., ACT-based policies~\cite{zhao2023learning}).
We treat these skills as controllers that share the same symbolic operators $\mathtt{Op}^\mathtt{C}$ but do not require separate samplers, as the low-level policy inherently proposes continuous actions until the desired effects are achieved.
Consequently, parametrized controllers with samplers and learned closed-loop skills are unified under a common operator-based abstraction for bilevel planning.

%% file: results.tex
\section{Experiment}
\label{sec:results}



\subsection{Implementation Details}
\myparagraph{Computational Hardware:}
Experiments for the \textit{Satellites}, \textit{Blocks}, \textit{Packing}, \textit{Table Clean Sim}, \textit{Table Clean Real} domains were conducted on a workstation equipped with a single NVIDIA RTX 4090 GPU and an Intel Ultra 265 CPU.
Conversely, the \textit{Tools} domain relies on point cloud observations; due to the higher computational demands of this observation space, we utilized a server featuring a single NVIDIA A100 GPU (80 GB VRAM) and an AMD EPYC 7543 32-Core CPU.

\begin{figure}[!t]
	\includegraphics[width=\columnwidth]{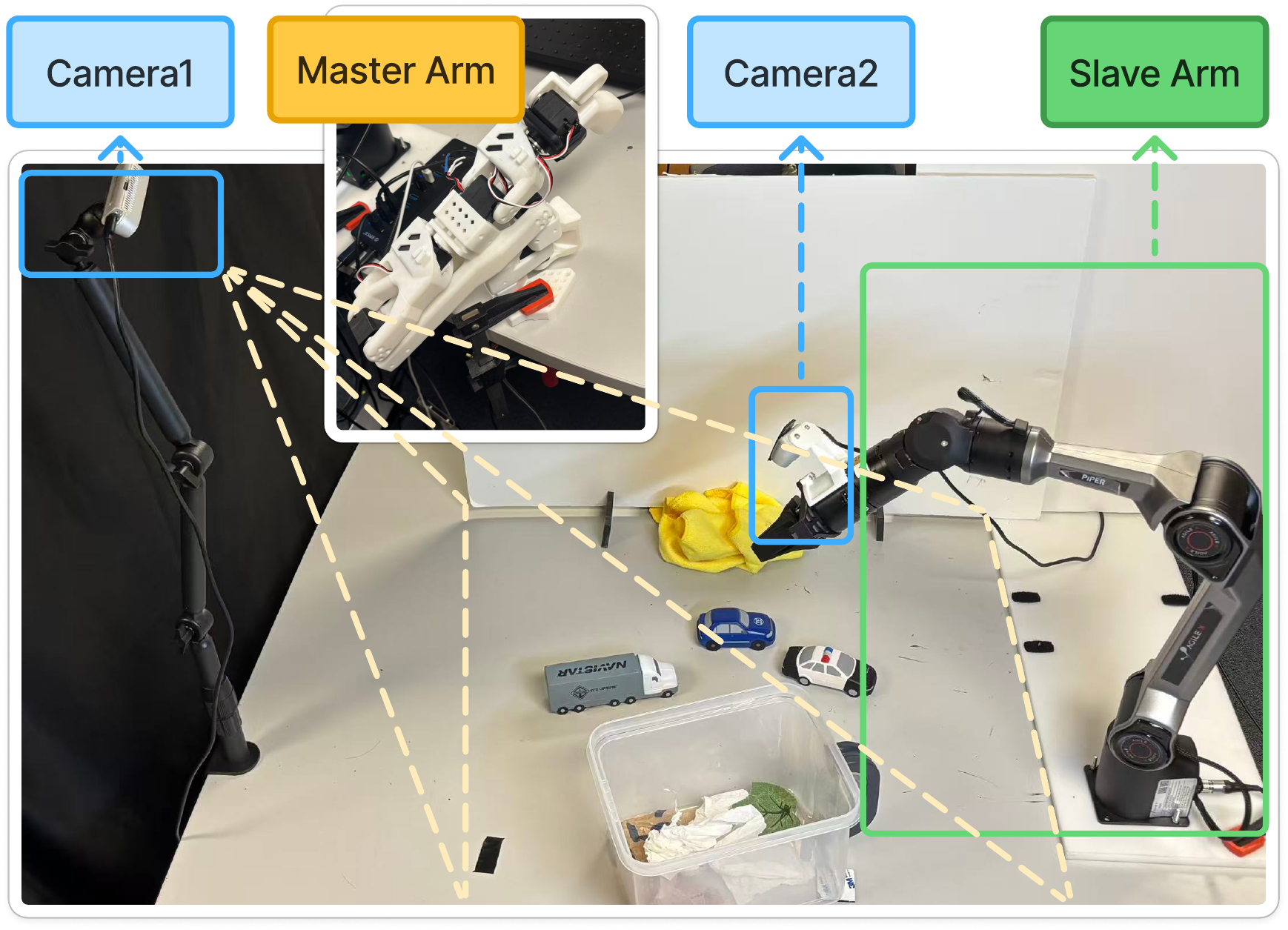}
	\caption{
Overview of our real-world experimental platform, showing the master arm used for teleoperation, the slave arm that executes the learned policies, and the two RGB cameras (Camera~1 and Camera~2) that capture observations of the table scene.
}

	\label{fig:real_system}
\end{figure}

\myparagraph{Real Robot Experiment Hardware:}
\label{sec:real_hardware}
As illustrated in Figure~\ref{fig:real_system}, our real-world experimental setup comprises an AgileX Robotics Piper 6-DoF manipulator and two Intel RealSense D435i cameras.
A static camera positioned frontally provides a global top-down view of the workspace, while a wrist-mounted camera captures egocentric visual data.
We collected demonstrations via the teleoperation interface depicted in the figure, using a smaller master arm to control the slave manipulator.
During data collection, we recorded the robot state and synchronized RGB streams from both cameras; these data serve as inputs for both predicate invention and the training of low-level controller skills.

\looseness = -1
\myparagraph{\shortname\ Training Details:}
{We employ \texttt{GPT\!-\!4o} as the default model for \shortname, and unless otherwise specified, all LLM- or VLM-based components throughout the system also rely on \texttt{GPT\!-\!4o}. For the baselines reported in Table~\ref{tab:results}, we additionally evaluate two model variants: \texttt{gemini-2.5-flash-lite} for the Gemini baseline and \texttt{qwen-plus} for the Qwen baseline. To ensure a fair comparison, all models operate at a fixed temperature of $0.2$. Each LLM call uses a structured prompt with four parts: (1) task and domain context, including the incomplete PDDL domain definition, object types, controller signatures, and a concise task description; (2) demonstration context, including operator demonstration sequences; (3) interaction history, including previously proposed predicate-effect hypotheses and their learning feedback when available; and (4) the requested output format, such as completing PDDL fragments or proposing predicate-effect hypotheses. These prompts include no hidden domain-specific hints beyond explicitly observable data.}

For predicate classifiers, we adopt architectures tailored to the observation modality of each domain. In domains where states are represented by continuous pose features, we train an {MLP encoder}. Conversely, for point-cloud observations, we employ a PointNet-based feature extractor. In image-based domains, visual embeddings are extracted using DINOv3 prior to classification. Across all domains, predicate classifiers are trained for a fixed number of epochs (typically 100). A predicate is designated as consistent when its classification loss converges below a specific threshold (typically 0.005). Further implementation details, hyperparameters, and training scripts are provided in our public code repository.

\myparagraph{Baselines:}
In simulated, non-image domains, we compare:
\begin{enumerate}
\item \textbf{IVNTR~\cite{li2025IVNTR}}: Neural loss guides symbolic effect search for predicate invention.
\item \textbf{GNN~\cite{battaglia2018gnn} and Transformer~\cite{vaswani2017tf}}: Relational policies trained by behavior cloning and evaluated with a shooting strategy.
\item \textbf{Grammar~\cite{silver2023predicateinvent}}: Predefined grammar rules propose predicates; this baseline fails to produce effective predicates in its three tested domains.
\item \textbf{Random}: Random plans provide a control baseline.
\end{enumerate}
In real-image domains, we compare:
\begin{enumerate}
\item \textbf{ViLa}: VLMs plan directly from images without learned symbolic abstractions. \textbf{ViLa-zero-shot}~\cite{hu2023look} uses the original prompts and in-context examples, with no demonstrations from our domains. \textbf{ViLa-few-shot}~\cite{athalye2024predicate} adds a small number of in-domain demonstrations. \textbf{ViLa-HPE} adds extensively hand-engineered constraints, such as requiring an empty gripper before picking. These constraints are not learned, and their manual specification limits scalability.
\item \textbf{Pix2Pred~\cite{athalye2024predicate}}: Given types $\Lambda$, a task description, and example images, a VLM proposes predicates and labels states; effect extraction and intersection produce operators without neural predicate classifiers. All variants use \texttt{GPT\!-\!4o}. \textbf{With GT predicates}, an oracle predicate set replaces invention while the VLM still labels atoms at inference. \textbf{With GT predicates and partial GT labels}, ground truth additionally replaces a subset of inference-time labels; the VLM labels the remaining atoms.
\item \textbf{ACT}~\cite{zhao2023learning}: Visuomotor imitation learns a separate policy for each skill, mapping images directly to low-level actions without symbolic abstractions.
\end{enumerate}

\myparagraph{Simulated Domains:}
Figure~\ref{fig:sim_domain} shows five domains with different state representations and generalization challenges:
\begin{itemize}
\item \textit{Satellites}~\cite{kumar2023predict}: Satellites collect sensor data from targets, represented by SE2 poses and object attributes. Training uses 2 satellites and 2 targets; testing uses 3 of each.
\item \textit{Blocks}~\cite{li2025IVNTR}: A robot arranges 3D blocks into specified two-level towers, effectively packing block pairs. Training uses 4--5 blocks; testing uses 6--7.
\item \textit{Tools}: Object-centric point clouds represent tools (wrench, screwdriver, hammer), fastening items (bolt, screw, nail), and contraptions. The goal is to secure each item to a designated board with the correct tool. Screwdriver--screw compatibility depends on geometry inferred from point clouds; wrench--bolt and hammer--nail matching depends only on type. The larger operator set expands the predicate search space. Training uses two items and two contraptions; testing uses two to three items and three contraptions.

\item \textit{Packing}: Boxes have adjustable dividers forming two regions, each holding at most one item. The robot must pack all differently sized items by selecting boxes and adjusting dividers to avoid overlap and respect capacity. States combine top-down images and object poses; region capacities are not directly available, so whether an item fits must be inferred visually. Test tasks contain more boxes and items than training tasks, requiring more coupled decisions and longer plans.

\item \textit{Table Clean Sim}: Our running example contains a robot, table, box, toys, towel, and dirt removable only by wiping. The goal is to store all toys in the box and remove all dirt. Wiping requires no toys on the table, to avoid sweeping them away, and the box away from the near side, to avoid collision; storing the towel requires the box on the near side for reachability. Object vectors contain 2D poses, cleanliness, and gripper status. Training tasks ($\mathcal{T}^\mathrm{train}$) start with exactly two toys on the table and an empty hand. Test tasks ($\mathcal{T}^\mathrm{test}$) contain three toys and unseen initial states, including an already grasped object or toys already in the box.

\end{itemize}

\myparagraph{Real Image Domains:}
Both domains use object-centric features from RGB observations:
\begin{itemize}
\item \textit{Table Clean Real}: The task and wiping/reachability constraints match \textit{Table Clean Sim}; RGB observations from the cameras in Section~\ref{sec:real_hardware} replace vector states with VFM embeddings of detected objects.
\item \textit{Cluttered Retrieval Real}: The robot retrieves targets into a box while reasoning about visual obstruction. It removes pickable obstacles (e.g., bottles) or repositions non-pickable ones (e.g., boxes). Training tasks contain two targets; test tasks contain three targets and more obstacles.
\end{itemize}

\myparagraph{Simulated Experiment Setup:}
An oracle bilevel planner collects demonstrations, and results are averaged over five random seeds. \textit{Satellites} and \textit{Blocks} use 500 trajectories per seed, following IVNTR. \textit{Tools} also uses 500 as a conservative, untuned budget; its classifiers and samplers are relatively easy to train once a valid abstraction is found, suggesting fewer demonstrations may suffice. \textit{Packing} and \textit{Table Clean Sim} use 50 trajectories per seed after verifying little impact on final planning performance, reducing data-generation cost. Each seed is evaluated on 50 tasks from $T \sim \mathcal{T}^\mathrm{train}$ and 50 from $T \sim \mathcal{T}^\mathrm{test}$ under a common maximum planning-time budget.

Total learning time is $\tau = \tau_{\mathrm{predinv}} + \tau_{\mathrm{predsel}} + \tau_{\mathrm{skill}}$: predicate invention (including LLM queries), predicate scoring/selection, and controller training. We measure these components separately for \shortname\ and use reported total training time for baselines with different pipelines.

\myparagraph{Real-robot Experiment Setup:}
\begin{figure}[!t]
	\includegraphics[width=\columnwidth]{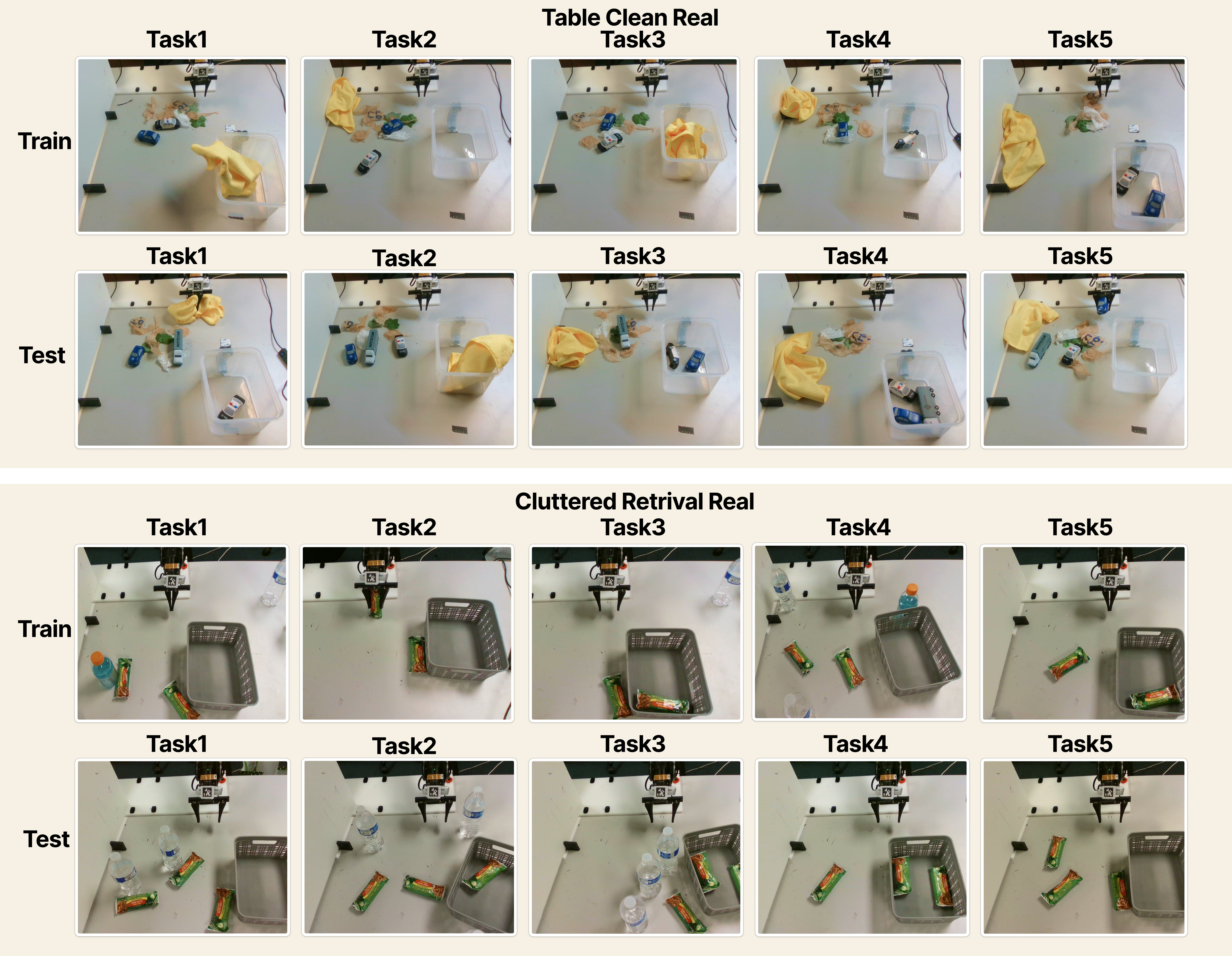}
	\caption{
{
Illustration of the ten tasks in the Table Clean Real and Cluttered Retrieval domains.
The top row shows the five training tasks, and the bottom row shows the five held-out test tasks.}
}
	\label{fig:table clean task}
\end{figure}
\looseness = -1
For each real domain, we collect 20 teleoperated demonstrations. Camera 1 RGB videos are manually segmented into actions, and representative state images are cropped using GroundingDINO~\cite{liu2023grounding}, with a custom DINO/SAM2 detector~\cite{ravi2024sam} for difficult objects. Analytic pick-and-place skills (e.g., \texttt{PickToyFromTable}) use SAM2 object point clouds to compute grasp/place poses. \texttt{PushBox} and \texttt{PullBox} use ACT trained on the corresponding 20 demonstrations, including master/slave arm states and both camera streams.

Offline image-based planning uses 10 tasks per domain (5 training, 5 testing; Figure~\ref{fig:table clean task}), with 10 random seeds per task and three runs per seed. We report average success on each split. Alongside the baselines above, we evaluate IVNTR with derived-aware selection. Because Pix2Pred struggles to invent \texttt{HandEmpty} and label relations such as $\texttt{Near}(\text{robot}, \text{box})$, its diagnostic variants supply oracle predicates and, for the partial-label variant, ground-truth $\texttt{Near(robot,box)}$ labels.

Physical execution uses the same 10 tasks per domain and three seeds per task, with up to three rollouts per seed. The fractional score is \(1/r\) for first success on rollout \(r\), or \(0\) if all fail. The human Oracle selects the next controller from the current state, using the same perception stack and low-level controllers as the autonomous methods.

\subsection{Empirical Results}
\label{sec:em_results}
\begin{figure*}[!t]
    \centering
    \includegraphics[width=2\columnwidth]{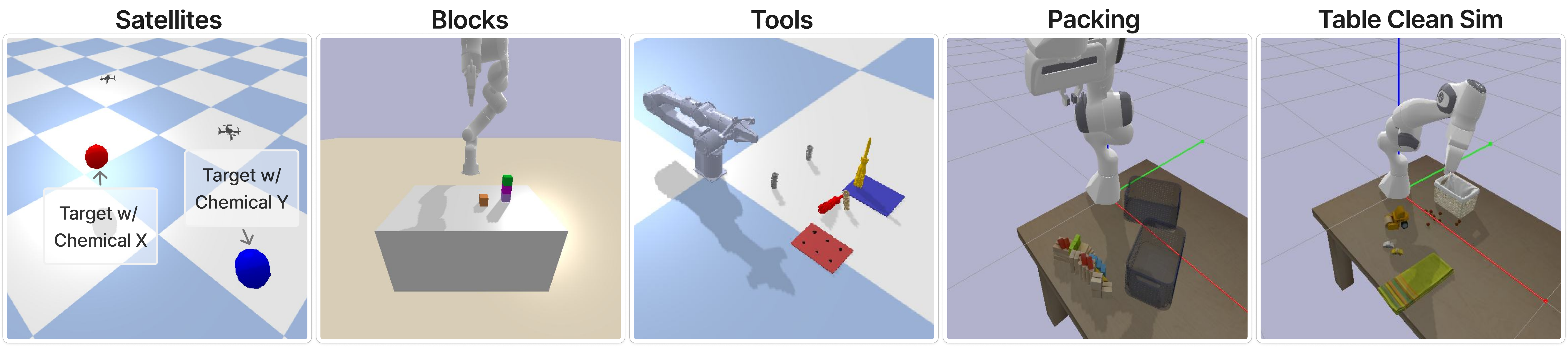}
    \caption{Visualization of the five simulated planning domains studied in this work.
    }
    \label{fig:sim_domain}
\end{figure*}

\myparagraph{Simulated domains.}
Table~\ref{tab:results} and Figure~\ref{fig:sim_domain} summarize the five domains. \shortname\ closely tracks the oracle on training tasks and generalizes to held-out configurations. It matches IVNTR's success in \textit{Satellites}, \textit{Blocks}, and \textit{Tools} while learning substantially faster, including \textbf{more than a factor of three speedup} in \textit{Tools}. Once a valid abstraction is found, \textit{Tools} reaches ceiling performance; its main distinction is learning efficiency. Standard IVNTR lacks the derived-aware selection needed for sound models in \textit{Packing} and \textit{Table Clean Sim}, so its success rates are not reported there; Section~\ref{sec:ablation} evaluates this module separately.

GNN and Transformer behavior-cloning policies can fit demonstrations in some domains but generalize poorly to more objects and longer horizons. Grammar-based invention is limited by state spaces that are difficult to express through predefined grammar functions.

Replacing the default language model with Qwen or Gemini retains high train/test success in the evaluated domains, with modest generalization losses and somewhat longer learning time (Table~\ref{tab:results}). Thus, the framework works with multiple backends, although model priors affect learning efficiency and test performance. Overall, \shortname\ combines the accuracy of relational predicate learning with lower training cost and stronger generalization than the behavior-cloning baselines.

\begin{table*}[ht]
\centering
\scriptsize
\setlength{\tabcolsep}{2pt}
\renewcommand{\arraystretch}{1.15}
\begin{adjustbox}{max width=\textwidth}
\begin{tabular}{l*{15}{c}}
\toprule
\multirow{2}{*}{\textbf{Method}} & 
\multicolumn{3}{c}{\textbf{Satellites (SE2)}} & 
\multicolumn{3}{c}{\textbf{Blocks (Vec3)}} & 
\multicolumn{3}{c}{\textbf{Tools (PCD)}} &
\multicolumn{3}{c}{\textbf{Packing (Image)}} &
\multicolumn{3}{c}{\textbf{Table Clean Sim (SE2)}} \\
\cmidrule(lr){2-4} \cmidrule(lr){5-7} \cmidrule(lr){8-10} \cmidrule(lr){11-13} \cmidrule(lr){14-16}
& $\mathcal{T}^{\text{train}}$ & $\mathcal{T}^{\text{test}}$ & $\tau$
& $\mathcal{T}^{\text{train}}$ & $\mathcal{T}^{\text{test}}$ & $\tau$
& $\mathcal{T}^{\text{train}}$ & $\mathcal{T}^{\text{test}}$ & $\tau$
& $\mathcal{T}^{\text{train}}$ & $\mathcal{T}^{\text{test}}$ & $\tau$
& $\mathcal{T}^{\text{train}}$ & $\mathcal{T}^{\text{test}}$ & $\tau$ \\
\midrule
Oracle          
& 100.0 & 100.0 & / 
& 100.0 & 100.0 & / 
& 100.0 & 100.0 & / 
& 100.0 & 100.0 & / 
& 100.0 & 100.0 & / \\

\textbf{UniPred (Ours)} 
& $\mathbf{99.6{\scriptstyle\pm0.9}}$ & $\mathbf{95.2{\scriptstyle\pm3.3}}$ & $\mathbf{1222.0{\scriptstyle\pm71.4}}$ 
& $\mathbf{99.2{\scriptstyle\pm0.8}}$ & $\mathbf{81.6{\scriptstyle\pm4.8}}$ & $\mathbf{4104.4{\scriptstyle\pm175.2}}$ 
& \textbf{100.0} & \textbf{100.0} & $\mathbf{11779.2{\scriptstyle\pm503.3}}$
& \textbf{100.0} & \textbf{100.0} & $\mathbf{3657.4{\scriptstyle\pm156.5}}$ 
& $\mathbf{96.0{\scriptstyle\pm1.6}}$ & $\mathbf{93.4{\scriptstyle\pm4.3}}$ & $\mathbf{764.6{\scriptstyle\pm32.9}}$  \\

IVNTR    
& $99.2{\scriptstyle\pm1.1}$ & $94.0{\scriptstyle\pm2.4}$ & $5129.8{\scriptstyle\pm405.0}$ 
& $99.2{\scriptstyle\pm0.8}$ & $73.2{\scriptstyle\pm4.4}$ & $12939.4{\scriptstyle\pm552.7}$  
& 100.0 & 100.0 & $45167.2{\scriptstyle\pm1929.4}$
& 0.0 & 0.0 & /
& 0.0 & 0.0 & / \\

GNN         
& $92.4{\scriptstyle\pm5.7}$  & $11.6{\scriptstyle\pm5.5}$  & $1303.0{\scriptstyle\pm35.8}$ 
& $89.2{\scriptstyle\pm4.4}$  & $38.8{\scriptstyle\pm6.5}$ & $1163.2{\scriptstyle\pm50.0}$ 
& 0.0   & 0.0  & $657.8{\scriptstyle\pm28.0}$ 
& 0.0  &  0.0 &  $1286.0{\scriptstyle\pm55.1}$ 
& 0.0  &  0.0 & $360.2{\scriptstyle\pm15.5}$\\

Transformer 
& $75.2{\scriptstyle\pm7.7}$ & $4.4{\scriptstyle\pm2.2}$  & $1778.6{\scriptstyle\pm96.0}$ 
& $34.4{\scriptstyle\pm6.2}$ & $16.8{\scriptstyle\pm3.2}$ & $1756.0{\scriptstyle\pm75.3}$ 
& 0.0  & 0.0  & $724.0{\scriptstyle\pm31.0}$
& 0.0 & 0.0 &  $1364.8{\scriptstyle\pm57.9}$
& 0.0 & 0.0 & $951.0{\scriptstyle\pm40.5}$\\

Grammar     
& 0.0 & 0.0 & / 
& 0.0 & 0.0 & / 
& /   & /   & /
& / & /& / 
& 0.0 & 0.0 & /\\

Random          
& 0.0  & 0.0  & / 
& $10.0{\scriptstyle\pm3.2}$ & $1.2{\scriptstyle\pm1.8}$  & / 
& 0.0  & 0.0  & /
& $12.4{\scriptstyle\pm3.4}$ & $1.6{\scriptstyle\pm2.1}$ & / 
& $3.6{\scriptstyle\pm3.0}$  & $1.2{\scriptstyle\pm1.8}$  & /\\

UniPred(Qwen)     
& $98.0{\scriptstyle\pm1.4}$ & $82.0{\scriptstyle\pm2.4}$ & $1493.0{\scriptstyle\pm118.6}$ 
& 100.0 & $75.6{\scriptstyle\pm5.2}$ & $5824.0{\scriptstyle\pm248.7}$  
& 100.0 & 100.0 & $15801.0{\scriptstyle\pm675.0}$
& 100.0 & 100.0 &  $4608.2{\scriptstyle\pm196.8}$
& $95.2{\scriptstyle\pm1.3}$ & $91.2{\scriptstyle\pm4.6}$  & $1274.6{\scriptstyle\pm54.6}$\\

UniPred(Gemini)  
& $99.2{\scriptstyle\pm1.1}$ & $94.0{\scriptstyle\pm1.4}$ & $2686.4{\scriptstyle\pm96.1}$
& 100.0 & $78.4{\scriptstyle\pm4.6}$ & $6894.6{\scriptstyle\pm294.8}$
& 100.0 & 100.0 & $14495.2{\scriptstyle\pm619.1}$
& 100.0 & 100.0 & $3937.0{\scriptstyle\pm168.4}$
& $96.4{\scriptstyle\pm1.5}$ & $90.4{\scriptstyle\pm4.9}$  & $883.0{\scriptstyle\pm38.0}$\\
\bottomrule
\end{tabular}
\end{adjustbox}
\caption{Success rate and runtime comparison on simulated domains. Values are reported as mean $\pm$ standard deviation over five seeds, with zero standard deviations omitted for compactness. $\mathcal{T}^{\text{train}}$ and $\mathcal{T}^{\text{test}}$ denote training and test distribution, respectively. $\tau$ is the total learning time (seconds). {Bold entries mark the default \shortname\ configuration for readability and should not be interpreted as statistical dominance on every metric.} Pix2Pred~\cite{athalye2024predicate} is not applicable here since these domains do not use RGB images only as state representations.}
\label{tab:results}
\end{table*}

\myparagraph{Task planning evaluation from images.}
Table~\ref{tab:real_image_task_eval} evaluates feasible plan generation from RGB images. \shortname\ achieves \(94.0\%\)/\(92.0\%\) train/test success in \textit{Table Clean Real} and \(94.0\%\)/\(98.0\%\) in \textit{Cluttered Retrieval Real}. IVNTR with derived-aware selection achieves comparable results (\(92.0\%\)/\(90.0\%\) and \(96.0\%\)/\(96.0\%\), respectively), demonstrating that this module transfers to other predicate-search methods for quantified conditions.

Without carefully engineered domain prompts, original Pix2Pred fails to propose key predicates such as \texttt{HandEmpty} and produces inconsistent labels in clutter, preventing reliable transition-model learning. This is a comparison of practical pixel-to-symbol grounding pipelines, not an isolated test of predicate-invention objectives: Pix2Pred uses VLM proposals and inference-time atom labeling, whereas \shortname\ trains classifier heads on detected object crops and DINO features from demonstrations.

Supplying oracle predicates alone yields Pix2Pred train/test success of \(47.3\%\)/\(33.3\%\) in table cleaning and \(22.0\%\)/\(18.0\%\) in cluttered retrieval. Adding partial ground-truth labels raises these to \(87.3\%\)/\(73.3\%\) and \(59.3\%\)/\(87.3\%\), still below \shortname. Remaining errors include confusing a target already \texttt{InBox} with one blocked by the box.

Direct VLM planners often ignore constraints or over-imitate demonstrations. In cluttered retrieval, ViLa-zero-shot and ViLa-few-shot achieve \(20.0\%\)/\(14.7\%\) and \(18.7\%\)/\(13.3\%\) train/test success; extensive prompt engineering raises test success to \(30.0\%\). Some methods score higher on test than training scenes despite more test objects: clearer layouts ease visual discrimination, and occasionally admit simple plans without enforcing every constraint (Figure~\ref{fig:table clean task}). This does not imply structurally easier test tasks. Learned classifiers also support domain adaptation through augmentation without new labeled training data, as evaluated in Section~\ref{sec:ablation}.

\par\smallskip
\myparagraph{Real Robot Execution.}
Table~\ref{tab:real_emp} reports fractional success under the protocol in Section~\ref{sec:real_hardware}. \shortname\ scores \(0.777\) in \textit{Table Clean Real} and \(0.867\) in \textit{Cluttered Retrieval Real}, exceeding Pix2Pred with GT predicates and partial GT labels (\(0.521\) and \(0.567\)). The Oracle reaches \(0.950\) in cluttered retrieval. Since \shortname\ shares its perception stack and controllers with the Oracle, both face the same perception and actuation limitations; \shortname's remaining failures mainly involve atom misclassification in states outside the training distribution, leading to incorrect controller choices.

ViLa-zero-shot, ViLa-few-shot, and ACT score zero in table cleaning; in cluttered retrieval they score \(0.189\), \(0.161\), and \(0.000\), respectively. Although they sometimes complete individual subtasks, they rarely maintain coherent long-horizon plans. Re-grounding after each controller and replanning when needed helps \shortname\ recover from execution deviations.

\myparagraph{Real world planning time.}
Across 100 images and three seeds, \shortname\ takes \(2.35\) seconds per query after object detection through complete plan generation: \(0.49\) seconds for DINO features and \(1.85\) seconds for grounding/planning. Direct GPT-4o planning takes \(4.02\) seconds zero-shot and \(9.31\) seconds few-shot (Table~\ref{tab:real_world_planning_time}).

\begin{table*}[!t]
    \centering
    \small
    \setlength{\tabcolsep}{6pt}
    \renewcommand{\arraystretch}{1.15}
    \begin{tabular*}{\textwidth}{@{\extracolsep{\fill}}lcccc@{}}
    \toprule
    \multirow{2}{*}{\textbf{Method}} & \multicolumn{2}{c}{\textbf{Table Clean Real}} & \multicolumn{2}{c}{\textbf{Cluttered Retrieval Real}} \\
    \cmidrule(lr){2-3} \cmidrule(lr){4-5}
    & $\mathcal{T}^{\text{train}}$ & $\mathcal{T}^{\text{test}}$ & $\mathcal{T}^{\text{train}}$ & $\mathcal{T}^{\text{test}}$ \\
    \midrule
    \textbf{\shortname\ (ours)} & $\mathbf{94.0 \pm 0.0}$ & $\mathbf{92.0 \pm 0.0}$ & $94.0 \pm 0.0$ & $\mathbf{98.0 \pm 0.0}$ \\
    IVNTR w/ derived & $92.0 \pm 0.0$ & $90.0 \pm 0.0$ & $\mathbf{96.0 \pm 0.0}$ & $96.0 \pm 0.0$ \\
    Pix2Pred w/ GT preds. and partial GT labels & $87.3 \pm 1.2$ & $73.3 \pm 2.3$ & $59.3 \pm 1.2$ & $87.3 \pm 1.2$ \\
    Pix2Pred w/ GT preds. & $47.3 \pm 3.1$ & $33.3 \pm 1.2$ & $22.0 \pm 3.5$ & $18.0 \pm 0.0$ \\
    ViLa-zero-shot & $3.3 \pm 1.2$ & $4.0 \pm 0.0$ & $20.0 \pm 2.0$ & $14.7 \pm 1.2$ \\
    ViLa-few-shot & $6.7 \pm 0.6$ & $0.0 \pm 0.0$ & $18.7 \pm 1.2$ & $13.3 \pm 3.1$ \\
    ViLa-HPE & $20.7 \pm 1.2$ & $11.3 \pm 1.2$ & $22.0 \pm 2.0$ & $30.0 \pm 0.0$ \\
    Pix2Pred & / & / & / & / \\
    \bottomrule
    \end{tabular*}
    \caption{Task planning success rates (\%) from real images in both domains. Values are mean $\pm$ standard deviation across three random seeds on training ($\mathcal{T}^{\text{train}}$) and held-out ($\mathcal{T}^{\text{test}}$) tasks. Bold marks the highest mean in each column.}
    \label{tab:real_image_task_eval}
\end{table*}
\begin{table}[!t]
    \centering
    \footnotesize
    \setlength{\tabcolsep}{3pt}
    \renewcommand{\arraystretch}{1.15}
    \begin{tabular*}{\columnwidth}{@{\extracolsep{\fill}}p{0.43\columnwidth}cc@{}}
    \toprule
    \textbf{Method} & \shortstack{\textbf{Table Clean}\\\textbf{Real}} & \shortstack{\textbf{Cluttered}\\\textbf{Retrieval Real}} \\
    \midrule
    Oracle (Human) & $0.872 \pm 0.095$ & $0.950 \pm 0.050$ \\
    \midrule
    \textbf{\shortname\ (ours)} & $\mathbf{0.777 \pm 0.048}$ & $\mathbf{0.867 \pm 0.029}$ \\
    \shortstack[l]{Pix2Pred w/ GT preds.\\and partial GT labels} & $0.521 \pm 0.139$ & $0.567 \pm 0.029$ \\
    ViLa-zero-shot & $0.000 \pm 0.000$ & $0.189 \pm 0.042$ \\
    ViLa-few-shot & $0.000 \pm 0.000$ & $0.161 \pm 0.019$ \\
    ACT & $0.000 \pm 0.000$ & $0.000 \pm 0.000$ \\
    \bottomrule
    \end{tabular*}
    \caption{Real-robot execution results. Values are mean $\pm$ sample standard deviation across three seeds, each averaged over ten tasks per domain (five training and five held-out tasks). A task scores $1/r$ if its first success occurs on rollout $r\leq 3$, and $0$ if all rollouts fail. Bold marks the highest mean among non-Oracle methods.}
    \label{tab:real_emp}
\end{table}

\begin{figure*}[!t]
	\includegraphics[width=2\columnwidth]{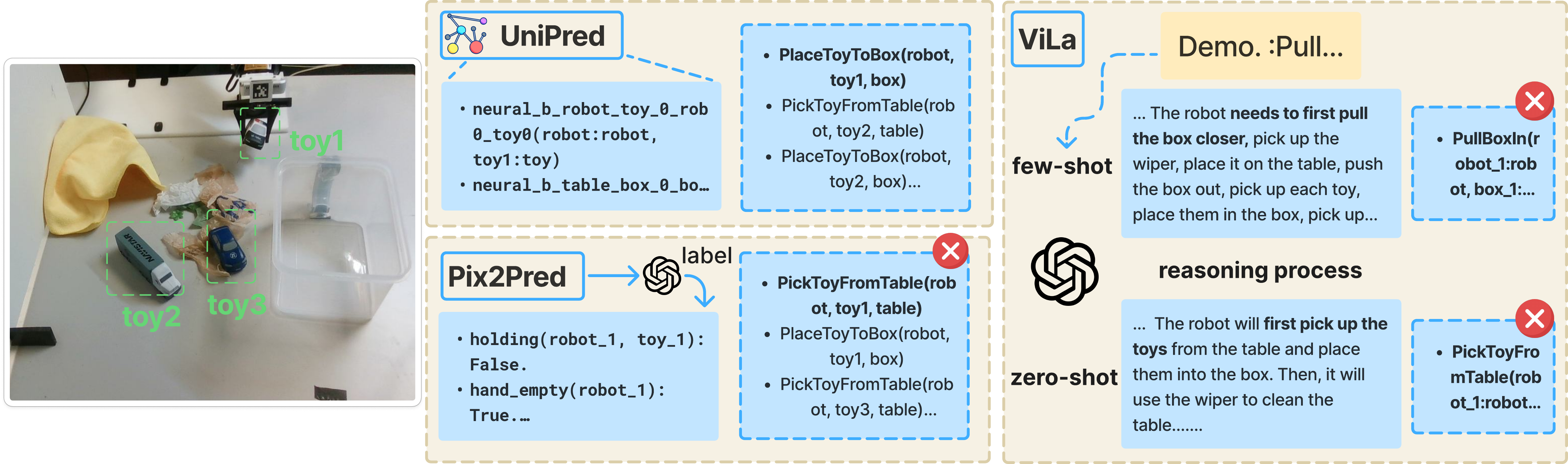}
	\caption{
Qualitative comparison on a real table-cleaning task. 
\shortname\ produces accurate symbolic predicates and action sequences, whereas Pix2Pred and ViLa generate incorrect or inconsistent next actions despite receiving the same visual observations.
}

	\label{fig:real_qual}
\end{figure*}

\subsection{Ablation Studies and Analysis}
\label{sec:ablation}
We now investigate the individual contributions of the primary components of \shortname\ through a series of ablation studies conducted in simulated domains.

\begin{figure}[!t]
	\includegraphics[width=\columnwidth]{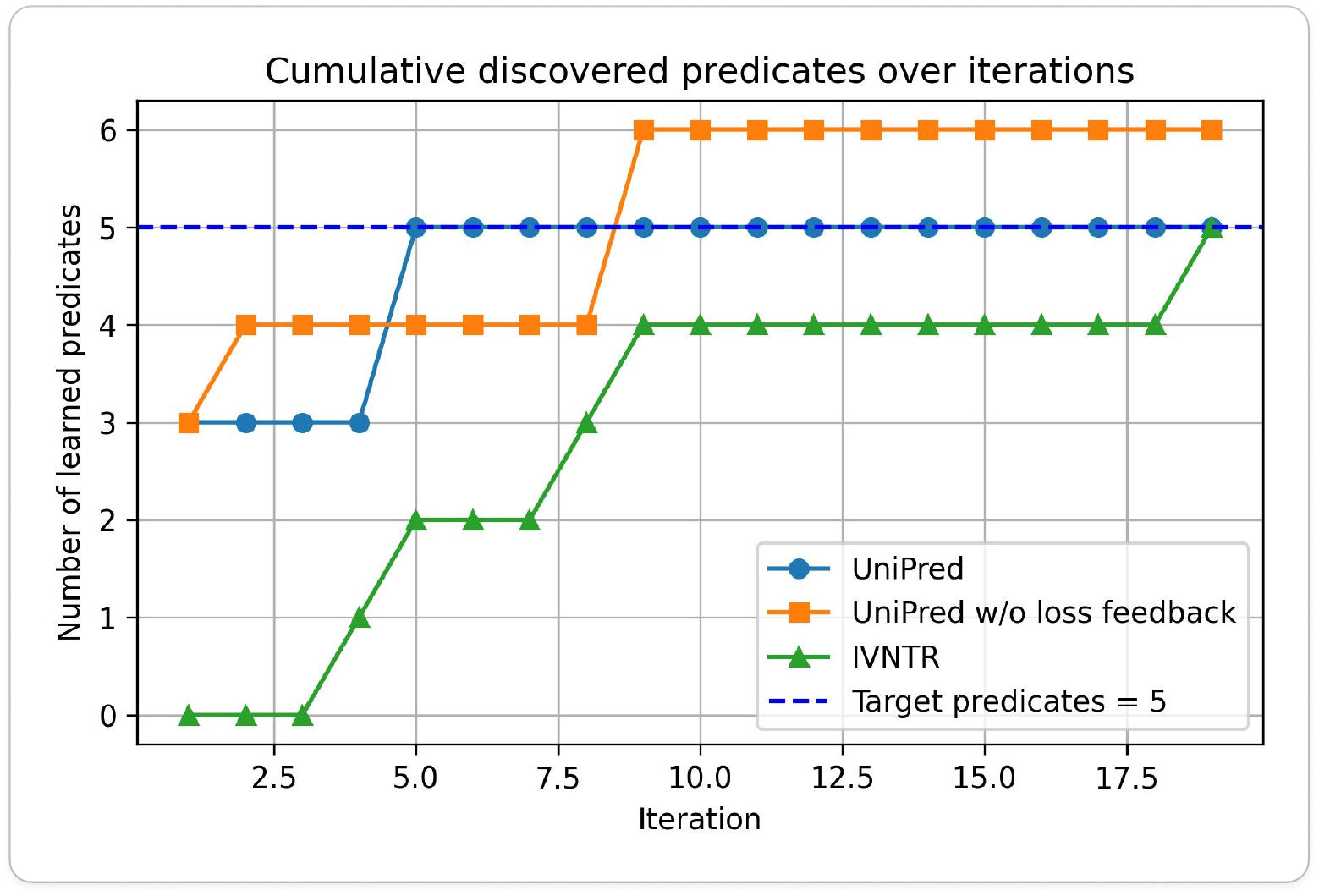}
	\caption{
Cumulative number of valid predicates discovered over iterations for UniPred, UniPred w/o loss feedback, and IVNTR, compared against the target set of five predicates (dashed line).
}

	\label{fig:learning_dynamic}
\end{figure}


\begin{table}[t]
    \centering
    \small
    \setlength{\tabcolsep}{4pt}
    \renewcommand{\arraystretch}{1.2}
    \begin{adjustbox}{max width=\columnwidth}
    \begin{tabular}{lccc}
    \toprule
    \textbf{Method} 
        & $\overline{\mathcal{T}^{\text{train}}}$ 
        & $\overline{\mathcal{T}^{\text{test}}}$ 
        & $\overline{t}$ \\
    \midrule
    \textbf{UniPred (Ours)}            
        & $\mathbf{99.7{\scriptstyle\pm0.2}}$
        & $\mathbf{94.2{\scriptstyle\pm1.2}}$
        & $\mathbf{5190.8{\scriptstyle\pm145.7}}$ \\
    UniPred w/o PDDL Completion        
        & $99.8{\scriptstyle\pm0.3}$
        & $90.0{\scriptstyle\pm7.8}$ 
        & $13665.5{\scriptstyle\pm610.9}$ \\
    UniPred w/o Unified Bilevel Learning       
        & $33.6{\scriptstyle\pm3.2}$ 
        & $25.0{\scriptstyle\pm1.7}$ 
        & $2870.7{\scriptstyle\pm58.7}$ \\
    UniPred w/o Loss Feedback          
        & $99.7{\scriptstyle\pm0.2}$ 
        & $94.4{\scriptstyle\pm0.9}$
        & $6006.4{\scriptstyle\pm144.4}$ \\
    \bottomrule
    \end{tabular}
    \end{adjustbox}
    \caption{Ablation result for the effect of unified bilevel learning in \shortname. Values are averaged over Satellites, Blocks, Tools, and Packing Sim, and reported as mean $\pm$ standard deviation over five seeds. {Bold entries mark the full \shortname\ configuration for readability and should not be interpreted as statistical dominance on every metric.}}
    \label{tab:ab_result_1}
\end{table}

\begin{table}[t]
    \centering
    \small
    \setlength{\tabcolsep}{4pt}
    \renewcommand{\arraystretch}{1.2}
    \begin{adjustbox}{max width=\columnwidth}
    \begin{tabular}{lcc}
    \toprule
    \multirow{2}{*}{\textbf{Method}} 
        & \multicolumn{2}{c}{\textbf{Blocks (Vec3)}} \\
    \cmidrule(lr){2-3}
        & $\mathcal{T}^{\text{train}}$ & $\mathcal{T}^{\text{test}}$ \\
    \midrule
    \textbf{UniPred (Ours)}        
        &$\mathbf{99.2{\scriptstyle\pm0.8}}$  
        &$\mathbf{81.6{\scriptstyle\pm4.8}}$  \\
    LLM complete PDDL               
        &$79.6{\scriptstyle\pm4.6}$  
        &$63.6{\scriptstyle\pm5.9}$  \\
    LLM propose code (Code as Policy~\cite{code_as_policy})
        &$12.4{\scriptstyle\pm6.7}$  
        &$1.2{\scriptstyle\pm1.8}$  \\
    \bottomrule
    \end{tabular}
    \end{adjustbox}
    \caption{
       Comparison between different ways of LLM proposing for planning. Values are reported as mean $\pm$ standard deviation over five seeds. 
    }
    \label{tab:ab_results_2}
\end{table}

\begin{table}[t]
    \centering
    \small
    \setlength{\tabcolsep}{4pt}
    \renewcommand{\arraystretch}{1.2}
    \begin{adjustbox}{max width=\columnwidth}
    \begin{tabular}{lccc}
    \toprule
    \textbf{Method} 
        & $\overline{\mathcal{T}^{\text{train}}}$ 
        & $\overline{\mathcal{T}^{\text{test}}}$ 
        & $\overline{t}$ \\
    \midrule
    \textbf{UniPred (Ours)}      
        & $\mathbf{99.6{\scriptstyle\pm0.3}}$ 
        & $\mathbf{92.3{\scriptstyle\pm1.6}}$ 
        & $\mathbf{5701.9{\scriptstyle\pm197.4}}$ \\
    UniPred w/o Semantic  
        & $100.0$
        & $91.9{\scriptstyle\pm1.8}$ 
        & $18612.9{\scriptstyle\pm558.2}$ \\
    IVNTR
        & $99.5{\scriptstyle\pm0.2}$
        & $89.1{\scriptstyle\pm2.0}$
        & $21078.8{\scriptstyle\pm754.6}$ \\
    \bottomrule
    \end{tabular}
    \end{adjustbox}
    \caption{
        Comparison between \shortname, \shortname\ w/o semantic information (removing operator names and names of known predicates from the prompt), and IVNTR, averaged over the three simulated domains. Values are reported as mean $\pm$ standard deviation over five seeds.
    }
    \label{tab:ab_results_3}
\end{table}
\begin{table}[t]
    \centering
    \small
    \setlength{\tabcolsep}{4pt}
    \renewcommand{\arraystretch}{1.2}
    \begin{adjustbox}{max width=\columnwidth}
    \begin{tabular}{lcc}
    \toprule
    \multirow{2}{*}{\textbf{Method}} 
        & \multicolumn{2}{c}{\textbf{Table Clean Sim (SE2)}} \\
    \cmidrule(lr){2-3}
        & $\mathcal{T}^{\text{train}}$ & $\mathcal{T}^{\text{test}}$ \\
    \midrule
    \textbf{UniPred (Ours)}              
        & $\mathbf{96.0{\scriptstyle\pm1.6}}$ 
        & $\mathbf{93.4{\scriptstyle\pm4.3}}$ \\
    IVNTR                                   
        & $0.0$  
        & $0.0$ \\
    UniPred w/o derived. select            
        & $0.0$  
        & $0.0$  \\
    IVNTR w/ derived. select          
        & $95.2{\scriptstyle\pm2.3}$  
        & $91.6{\scriptstyle\pm3.0}$ \\
    \bottomrule
    \end{tabular}
    \end{adjustbox}
    \caption{Ablation study for the derived-aware predicate selection module. Values are reported as mean $\pm$ standard deviation over five seeds; zero standard deviations are omitted for compactness.}
    \label{tab:ab_result_4}
\end{table}

\begin{table*}[t]
    \centering
    \small
    \setlength{\tabcolsep}{4pt}
    \renewcommand{\arraystretch}{1.15}
    \begin{adjustbox}{max width=\textwidth}
    \begin{tabular}{lcccccccc}
    \toprule
    \multirow{2}{*}{\textbf{Domain}}
        & \multicolumn{6}{c}{\textbf{Time Breakdown (s)}}
        & \multicolumn{2}{c}{\textbf{LLM Usage}} \\
    \cmidrule(lr){2-7} \cmidrule(lr){8-9}
        & \textbf{Total}
        & \textbf{Classifier Training}
        & \textbf{LLM Query}
        & \textbf{Pred. Selection}
        & \textbf{Sampler Learning}
        & \textbf{Rest}
        & \textbf{Tokens}
        & \textbf{Cost (\$)} \\
    \midrule
    Satellites
        & $1232.3 \pm 71.7$
        & $391.3 \pm 32.7$
        & $126.7 \pm 46.9$
        & $469.3 \pm 24.0$
        & $75.7 \pm 16.3$
        & $169.3 \pm 24.4$
        & $44823.7 \pm 16625.6$
        & $0.147 \pm 0.053$ \\

    Blocks
        & $4182.0 \pm 268.5$
        & $520.0 \pm 29.7$
        & $144.0 \pm 68.8$
        & $2921.3 \pm 226.4$
        & $442.0 \pm 34.1$
        & $154.7 \pm 37.2$
        & $53734.0 \pm 32245.1$
        & $0.166 \pm 0.095$ \\
    \bottomrule
    \end{tabular}
    \end{adjustbox}
    \caption{
        {UniPred learning time breakdown and LLM usage across domains.
        Each entry reports the mean and standard deviation across three random seeds.
        Time is measured in seconds.}
    }
    \label{tab:unipred_learning_time_breakdown}
\end{table*}
\begin{table}[t]
    \centering
    \small
    \setlength{\tabcolsep}{6pt}
    \renewcommand{\arraystretch}{1.2}
    \begin{adjustbox}{max width=\columnwidth}
    \begin{tabular}{lccc}
    \toprule
    \multirow{2}{*}{\textbf{Method}}
        & \multirow{2}{*}{\textbf{Planning Time per Query (s)}}
        & \multicolumn{2}{c}{\textbf{\shortname Time Breakdown (s)}} \\
    \cmidrule(lr){3-4}
        &
        & \textbf{Perception}
        & \textbf{Planning} \\
    \midrule
    \textbf{\shortname (Ours)}
        & $\mathbf{2.35 \pm 0.91}$
        & $0.49$
        & $1.85$ \\
    ViLa-zero-shot
        & $4.02 \pm 0.82$
        & N/A
        & N/A \\
    ViLa-few-shot
        & $9.31 \pm 2.41$
        & N/A
        & N/A \\
    \bottomrule
    \end{tabular}
    \end{adjustbox}
    \caption{
        {Real-world planning time per query on 100 images over three seeds. Timing excludes object detection and ends when the plan is generated. For \shortname, perception denotes DINO feature extraction and planning denotes predicate grounding plus plan generation.}
    }
    \label{tab:real_world_planning_time}
\end{table}

\looseness = -1
\myparagraph{Effect of unified bilevel learning.}
Table~\ref{tab:ab_result_1} averages train/test success and total learning time across \textit{Satellites}, \textit{Blocks}, \textit{Tools}, and \textit{Packing Sim}, using five seeds. The full pipeline combines PDDL initialization, iterative bilevel learning, and loss feedback. ``W/o PDDL Completion'' samples initial candidates from the LLM without PDDL warm-starting. ``W/o Unified Bilevel Learning'' uses a single PDDL completion without refinement. ``W/o Loss Feedback'' retains the loop and candidate acceptance/rejection signals but removes the scalar \([0,100]\) loss score, relying on language priors and simple heuristics.

The full method achieves \(99.7\%\)/\(94.2\%\) train/test success in \(5190.8\) seconds. Removing PDDL initialization increases learning time to \(13665.5\) seconds and reduces test success to \(90.0\%\). A single PDDL completion takes only \(2870.7\) seconds but yields \(33.6\%\)/\(25.0\%\) success. Without scalar loss feedback, accuracy remains comparable (\(99.7\%\)/\(94.4\%\)), while learning takes \(6006.4\) seconds. Loss feedback thus improves search efficiency rather than being necessary for high final accuracy. Figure~\ref{fig:learning_dynamic} compares predicate discovery across iterations. Both \shortname\ variants use language priors to propose early candidates satisfying most structural constraints, while IVNTR explores substantially more candidates. Scalar loss feedback gives the LLM finer guidance than acceptance/rejection alone, helping refine the proposals and reducing the search needed to obtain useful predicates. This behavior is consistent with the learning-time comparison in Table~\ref{tab:ab_result_1}; it does not imply that scalar feedback is required to reach the final success rate.

\myparagraph{Effect of different LLM interfaces.}
We subsequently evaluate alternative strategies for utilizing the LLM within the \textit{Blocks} domain, as summarized in \tref{tab:ab_results_2}. 
Our proposed approach, which employs the LLM solely to propose predicate templates that are subsequently trained and evaluated via the bilevel pipeline, achieves a test success rate of \(81.6\%\). 
Conversely, restricting the LLM to generate a static PDDL domain description, denoted as ``LLM complete PDDL'', reduces test success to \(63.6\%\). 
Although the generated PDDL is typically syntactically valid, it frequently omits critical preconditions or effects and fails to adapt adequately to the statistical distribution of the demonstration data.

The most direct strategy, ``LLM propose code (Code as Policy~\cite{code_as_policy})'', prompts the LLM to output executable planning code that operates on low level states and selects primitive actions. 
{
Because this ablation is in the simulated \textit{Blocks} domain, we evaluate the generated code using ground truth simulator states rather than image perception. 
The LLM is given the environment interface, including object states, actions, and a language description of the task goal and constraints. 
We execute the generated code on the same train and test tasks and report final task success.
}

This variant achieves only \(1.2\%\) test success despite somewhat higher training success. In this setting, using the LLM to guide data-grounded predicate search is more effective than directly generating complete planning programs.

\myparagraph{Role of semantic information in the LLM prompt.}
The ``w/o Semantic'' variant removes operator and known-predicate names, retaining type signatures and a minimal task description. Table~\ref{tab:ab_results_3} compares it with \shortname\ and IVNTR under the same protocol. Removing names maintains high accuracy (\(100.0\%\)/\(91.9\%\) train/test, versus \(99.6\%\)/\(92.3\%\) for the full method) but increases learning time from \(5.7 \times 10^3\) to \(1.86 \times 10^4\) seconds. IVNTR achieves \(99.5\%\)/\(89.1\%\) in \(2.11 \times 10^4\) seconds. Semantic context primarily reduces exploration cost; even without it, the LLM-guided loop remains more efficient than IVNTR's tree expansion.

\begin{figure}[!t]
	\includegraphics[width=\columnwidth]{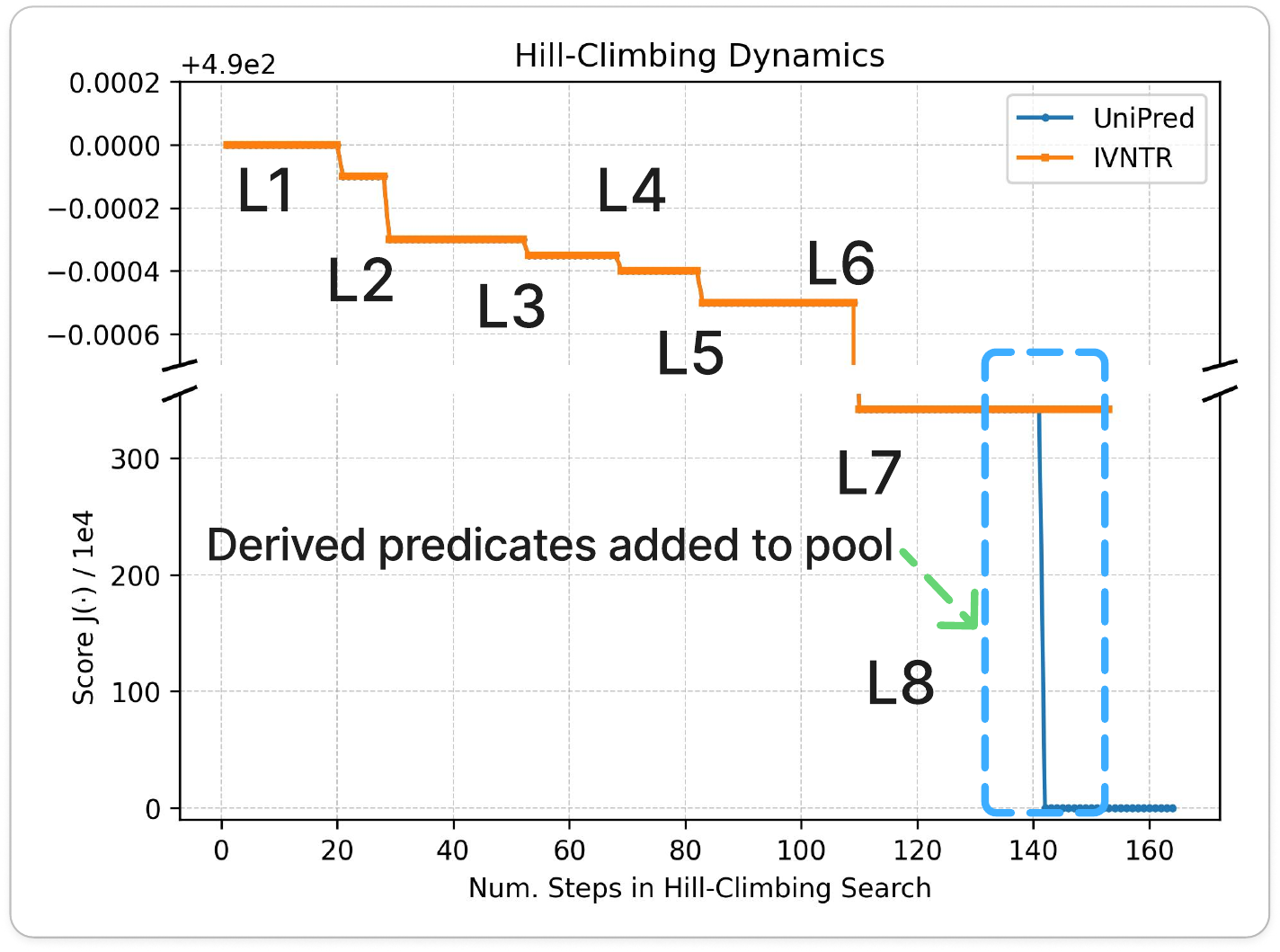}
	\caption{
Hill-climbing dynamics of the predicate selection score for IVNTR and UniPred  
The curve shows how the score changes over hill-climbing steps: most updates cause only small score adjustments that are visually subtle, while a few iterations produce noticeable drops when useful derived predicates are added to the pool.
}

	\label{fig:hill_climbing}
\end{figure}

\myparagraph{Derived-aware predicate selection.}
\label{sec:derived_ab}
In \textit{Table Clean Sim}, wiping requires quantified conditions such as all toys being placed. Both \shortname\ and IVNTR achieve high success with derived-aware selection but \(0.0\%\) on both splits without it (Table~\ref{tab:ab_result_4}). The module's benefit is therefore transferable, rather than exclusive to \shortname. Without it, treating basic and derived predicates alike prevents discovery of the required quantified condition, and the hill-climbing score plateaus. With it, adding a useful derived predicate sharply reduces the score (Figure~\ref{fig:hill_climbing}).

\myparagraph{Learning time breakdown and LLM usage.}
Table~\ref{tab:unipred_learning_time_breakdown} reports reference costs. Total learning takes \(1232.3\) seconds in \textit{Satellites} and \(4182.0\) seconds in \textit{Blocks}; predicate selection and classifier training dominate, with more costly selection in \textit{Blocks}. LLM queries take \(126.7\) and \(144.0\) seconds, respectively. Average usage is \(44823.7\) tokens (\(\$0.147\)) in \textit{Satellites} and \(53734.0\) tokens (\(\$0.166\)) in \textit{Blocks}.

\begin{figure}[!t]
	\includegraphics[width=\columnwidth]{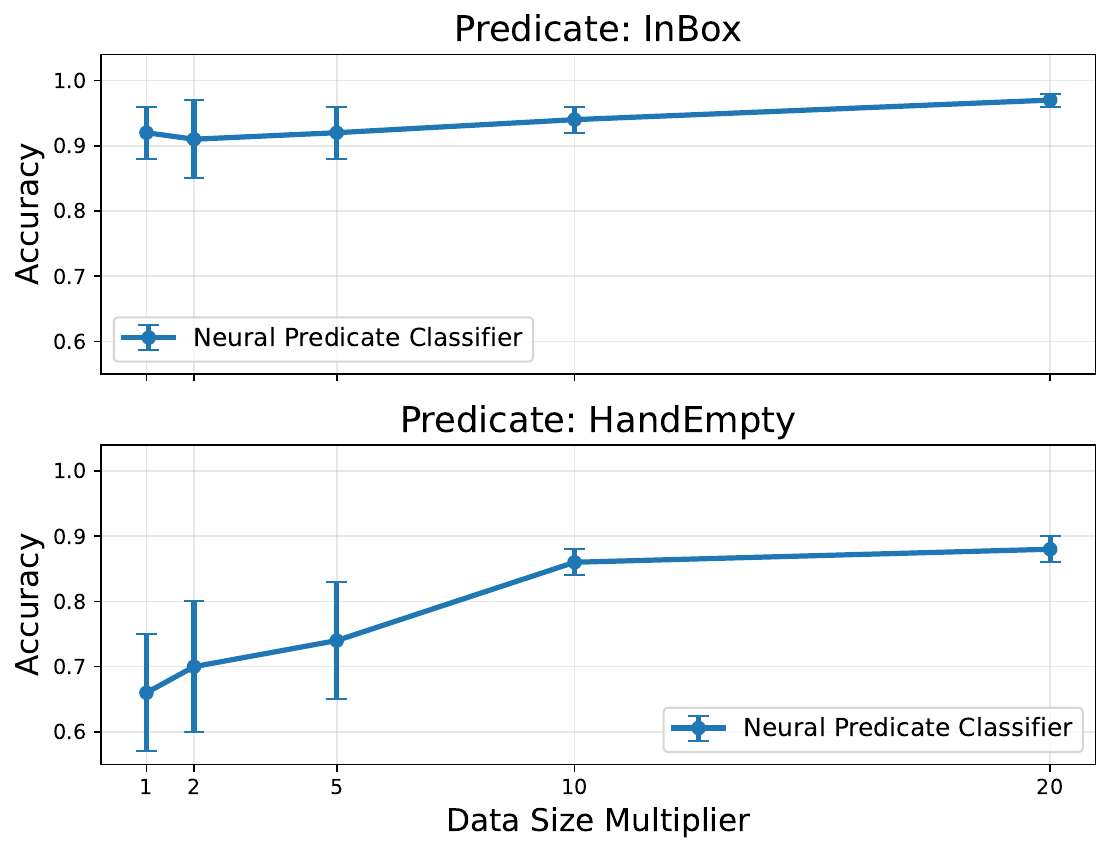}
	\caption{
{Out-of-distribution predicate-classifier accuracy for \textit{InBox} and \textit{HandEmpty} on 100 real images with novel objects/backgrounds. The x-axis is the augmentation multiplier applied to the original 20-demonstration training set.}
}

	\label{fig:pred_size}
\end{figure}

\myparagraph{Robustness of DINO-based predicate classifier.}
We evaluate \textit{InBox} and \textit{HandEmpty} on novel objects and backgrounds in \textit{Table Clean}, using 100 real images per predicate (50 positive, 50 negative). Classifiers trained on the original demonstrations have limited robustness under this severe visual shift, especially for \textit{HandEmpty}. Augmenting the original demonstrations improves accuracy \textbf{without additional labeled training data}: a \(20\times\) augmented set yields \(97\%\) accuracy for \textit{InBox} and \(88\%\) for \textit{HandEmpty} (Figure~\ref{fig:pred_size}).

\begin{figure}[!t]

    \includegraphics[width=\columnwidth]{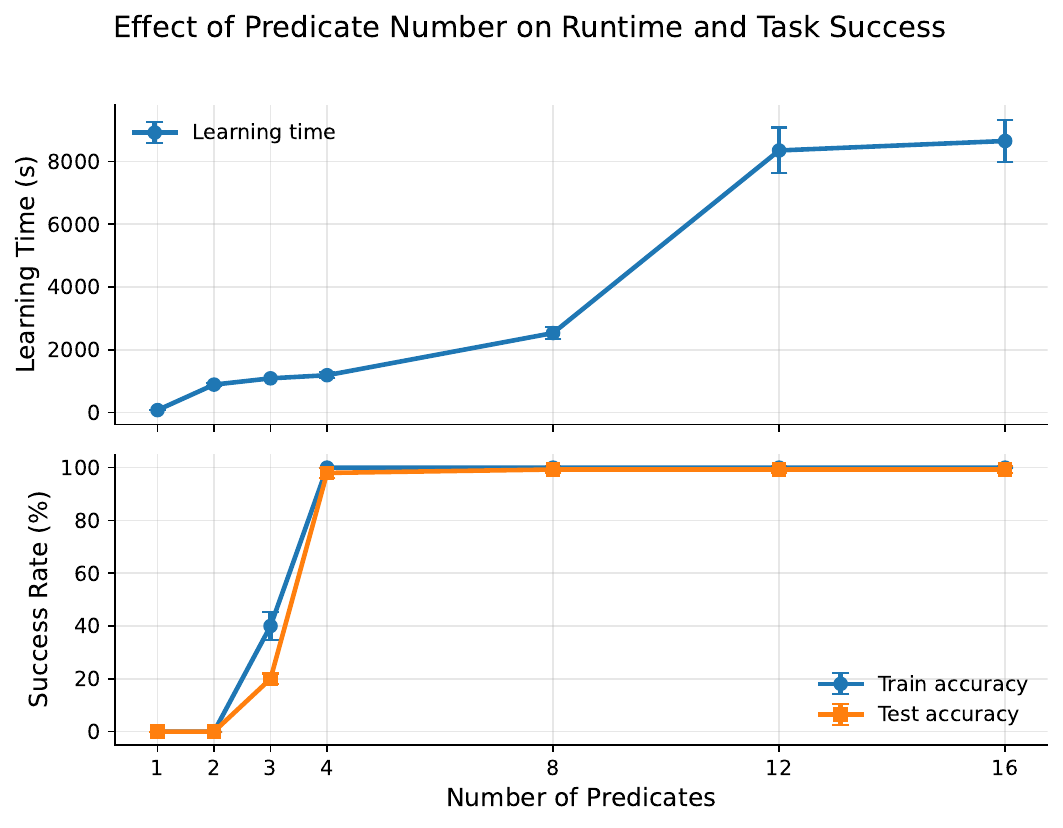}
\caption{
{Effect of the hyperparameter specifying the expected number of basic predicates on learning time and task success rate in the \textit{Satellites} domain. Results are averaged over three random seeds, with error bars showing standard deviation.}
}
	\label{fig:pred_num}
\end{figure}

\myparagraph{Effect of the expected predicate count.}
Figure~\ref{fig:pred_num} varies the expected number of accepted basic predicates in \textit{Satellites}, averaging three seeds. Too few predicates prevent successful planning; at four predicates, train/test success approaches \(100\%\). Larger budgets add learning cost without improving success, since further candidates must satisfy the loss criterion even when unnecessary for planning. Runtime growth moderates after most acceptable predicates are found. Performance is thus stable above the minimum sufficient count, while excessive budgets waste search effort.


%% file: conclusion.tex
\section{Discussions}
\label{sec:conclusion}
\myparagraph{Limitations and Future Work:} Despite its effectiveness in long-horizon planning, our \shortname\ framework has several limitations that suggest promising avenues for future research.
(1) Similar to prior work such as IVNTR \cite{li2025IVNTR}, our predicate classifiers are supervised with effect vectors. This imposes the implicit constraint that each low-level controller must correspond to a unique, deterministic STRIPS operator in the symbolic domain (with finite object-centric effects). 
A crucial future direction is investigating how symbolic predicates can be discovered in contexts involving non-STRIPS controllers (with continuous, non-deterministic action effects)~\cite{llm4pddlstream,mao2024hybrid}.
(2) Our predicate invention framework currently requires a fixed, provided set of low-level controllers, which are typically learned via imitation. 
{Predicate quality is therefore relative to this controller set.}
Future work should explore extending the framework to jointly discover new operational skills~\cite{liu2025slap,akhil2021dsg,eysenbach2019dyna} and the corresponding high-level predicates and symbolic abstractions that govern them.
(3) Since \shortname\ learns its abstractions directly from human demonstrations and indirectly through the representations of foundation models, the discovered predicates are inherently constrained by human prior knowledge. 
Discovering emergent, non-intuitive, or ``surprisingly smart" abstractions~\cite{liu2025slap} that move beyond human preconceptions remains an intriguing and important direction.
(4) Following established works in bilevel planning \cite{chitnis2021nsrt,li2025IVNTR,silver2023predicateinvent}, \shortname\ operates under the assumption of a fully-observable and deterministic environment. A necessary next step is studying how to robustly learn and utilize these symbolic predicates within more challenging contexts, such as Partially Observable Markov Decision Processes (POMDPs) and environments characterized by stochastic physics.
{(5) Since \shortname\ already combines LLM-guided symbolic proposal with learned perceptual predicate grounding, it is naturally compatible with agentic robot frameworks, where coding agents call perception and action tools to solve manipulation tasks~\cite{fu2025capx}. A key open problem for such agents is how to build task-relevant domain representations rather than only compose tool calls. Learning neural-symbolic world models from data, as in \shortname, offers one possible direction.}

\myparagraph{Conclusion:}
We presented \shortname, a unified bilevel framework for discovering neural-grounded symbolic predicates. It combines LLM-guided proposals with bottom-up learning feedback that refines predicate hypotheses using low-level data, derived-aware predicate selection for Non-STRIPS domains, and classifier optimization on visual foundation-model features~\cite{simeoni2025dinov3} for perceptual generalization from few real-world demonstrations. Across five simulated and two real-world domains with diverse state representations, \shortname\ achieves $2\sim4\times$ higher success rates than purely top-down methods and $3\sim4\times$ faster learning than purely bottom-up approaches. These results demonstrate the benefits of combining foundation-model priors with data-grounded abstraction learning for efficient, generalizable long-horizon robot planning.

%% file: acknowledgment.tex
\section*{Acknowledgment}
This work has been funded in part by the Army Research Laboratory (ARL) award W911NF-23-2-0007 and W911QX-24-F-0049, and the Office of Naval Research (ONR) award N00014-23-1-2840.
We also acknowledge the support of the Air Force Research Laboratory (AFRL), the Defense Advanced Research Projects Agency (DARPA), under agreement number FA8750-23-2-1015, and the Defence Science and Technology Agency (DSTA) under contract \#DST000EC124000205.
This work used Bridge s-2 at PSC through allocation cis220039p from the Advanced Cyberinfrastructure Coordination Ecosystem: Services \& Support (ACCESS) program which is supported by NSF grants \#2138259, \#2138286, \#2138307, \#2137603, and \#213296.

%% file: biographies.tex
\begingroup
\makeatletter
\def\@IEEEBIOphotowidth{1.0in}
\def\@IEEEBIOphotodepth{1.25in}
\def\@IEEEBIOhangwidth{1.14in}
\def\@IEEEBIOhangdepth{1.25in}
\def\@IEEEBIOskipN{6pt}
\makeatother
\newsavebox{\biographyphotobox}
\newlength{\biographyphototrim}
\newcommand{\biographyphoto}[1]{%
  \sbox{\biographyphotobox}{\includegraphics[height=1.25in]{imgs/authors/#1}}%
  \parbox[c][1.25in][t]{1.0in}{%
    \ifdim\wd\biographyphotobox<1.0in\relax
      \sbox{\biographyphotobox}{\includegraphics[width=1.0in]{imgs/authors/#1}}%
      \setlength{\biographyphototrim}{\dimexpr\ht\biographyphotobox-1.25in\relax}%
      \clipbox{0pt \biographyphototrim{} 0pt 0pt}{\usebox{\biographyphotobox}}%
    \else
      \setlength{\biographyphototrim}{\dimexpr(\wd\biographyphotobox-1.0in)/2\relax}%
      \clipbox{\biographyphototrim{} 0pt \biographyphototrim{} 0pt}{\usebox{\biographyphotobox}}%
    \fi
  }%
}

\Needspace{1.45in}
\begin{IEEEbiography}[{\biographyphoto{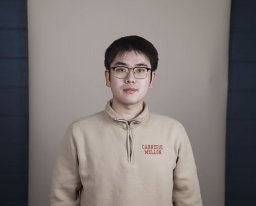}}]{Qianwei Wang}
received the B.S. degree from the University of Michigan,
Ann Arbor, MI, USA. He is currently pursuing the Ph.D. degree in
electrical and computer engineering at Princeton University,
Princeton, NJ, USA, where he is advised by Tom Silver. His research interests include
physical reasoning for robots, long horizon planning, contact rich
manipulation, and robot learning.
\end{IEEEbiography}

\Needspace{1.45in}
\begin{IEEEbiography}[{\biographyphoto{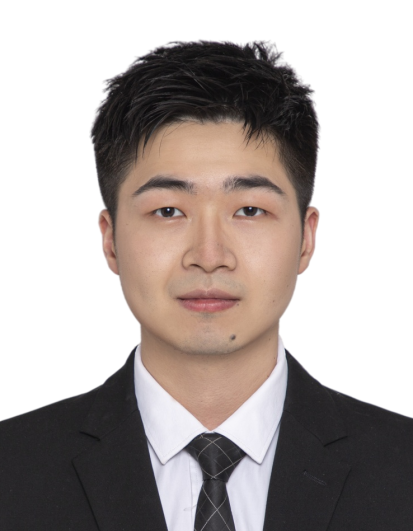}}]{Bowen Li}
received the B.Eng. degree in mechanical engineering from
Tongji University, Shanghai, China, in 2022, and the M.S. degree in
robotics from Carnegie Mellon University, Pittsburgh, PA, USA, in
2024. He is currently pursuing the Ph.D. degree in robotics at
Carnegie Mellon University, advised by Sebastian Scherer. His research
focuses on long horizon embodied reasoning, robot program synthesis,
reinforcement learning, and planning.
\end{IEEEbiography}

\Needspace{1.45in}
\begin{IEEEbiography}[{\biographyphoto{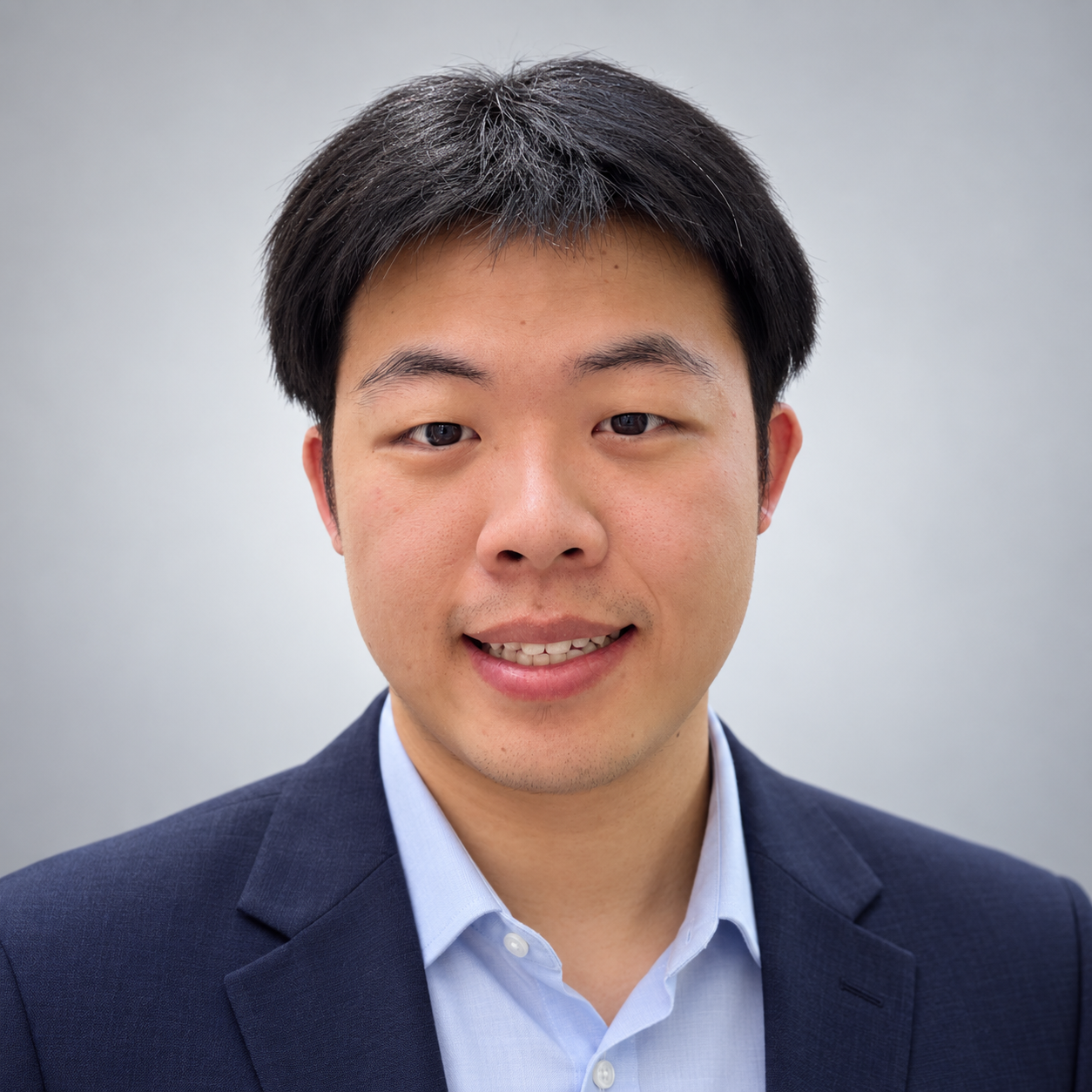}}]{Zhanpeng Luo}
received the bachelor's degree in computer science from
the University of Pittsburgh, Pittsburgh, PA, USA. He is currently
pursuing the Ph.D. degree in computer science at Johns Hopkins
University, Baltimore, MD, USA, advised by Jaemin Cho. His research
interests include generative modeling, multimodal learning, and
learning actions from unlabeled data.
\end{IEEEbiography}

\Needspace{1.45in}
\begin{IEEEbiography}[{\biographyphoto{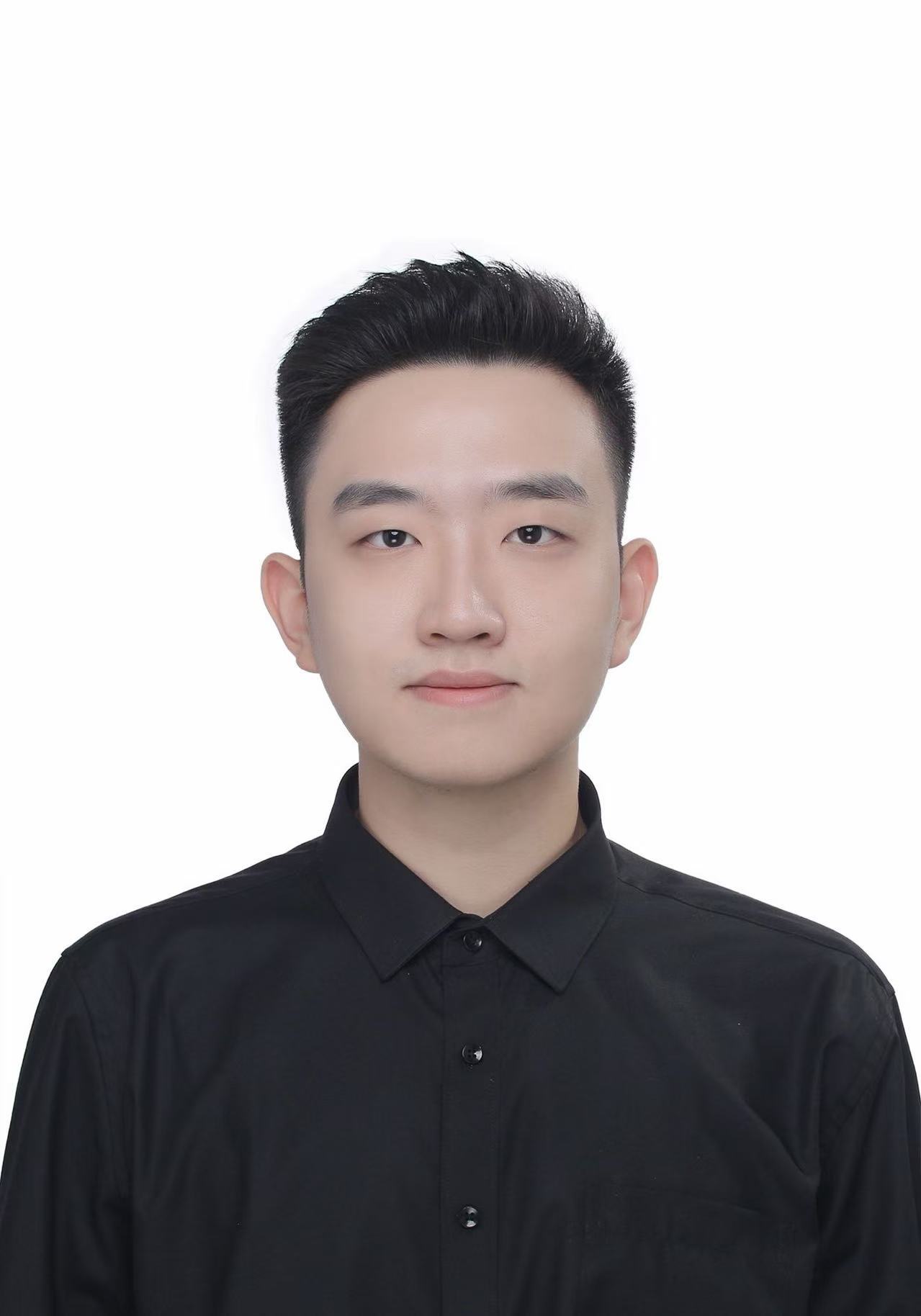}}]{Yifan Xu}
received the B.S.E. degree in robotics from Northeastern
University, Shenyang, China, in 2021, the M.S. degree in robotics
from the University of Michigan, Ann Arbor, MI, USA, in 2023, and
the Ph.D. degree in civil and environmental engineering from the
University of Michigan in 2026. He is currently a Research Assistant
Professor at the University of Tennessee Oak Ridge Innovation
Institute, Oak Ridge, TN, USA. His research interests include
construction robotics, assistive robotics, human robot collaboration,
and multimodal artificial intelligence for robotic scene understanding.
\end{IEEEbiography}

\Needspace{1.45in}
\begin{IEEEbiography}[{\biographyphoto{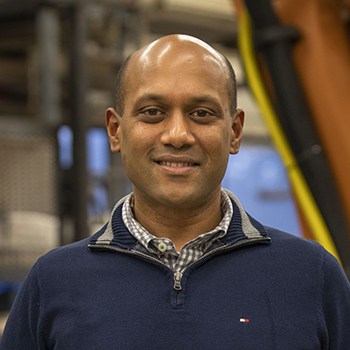}}]{Vineet Kamat}
received the B.E. degree in civil engineering from
Goa University, Goa, India, in 1998, and the M.S. and Ph.D. degrees
in civil engineering from Virginia Tech, Blacksburg, VA, USA, in
2000 and 2003, respectively. He is the John L. Tishman Family
Professor of Construction Management and Sustainability at the
University of Michigan, Ann Arbor, MI, USA. His research focuses on
automation and robotics for the construction, operation, and
maintenance of civil infrastructure and the builtCamera environment.
He is a Fellow of ASCE.
\end{IEEEbiography}

\Needspace{1.45in}
\begin{IEEEbiography}[{\biographyphoto{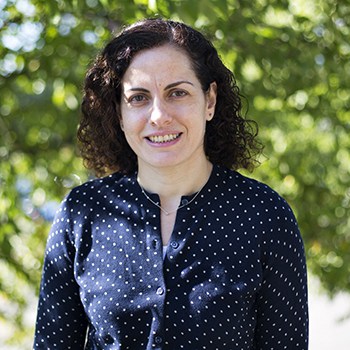}}]{Carol Menassa}
received the B.S. and M.S. degrees in civil and
environmental engineering from the American University of Beirut,
Beirut, Lebanon, in 1995 and 2002, respectively, and the M.S. degree
in finance and Ph.D. degree in civil and environmental engineering
from the University of Illinois Urbana Champaign, Champaign, IL,
USA, in 2009. She is the Robert I. Carr Collegiate Professor of Civil
and Environmental Engineering at the University of Michigan,
Ann Arbor, MI, USA. Her research focuses on human centered automation,
robotics, and intelligent built environments. She is a Fellow of ASCE.
\end{IEEEbiography}

\Needspace{1.45in}
\begin{IEEEbiography}[{\biographyphoto{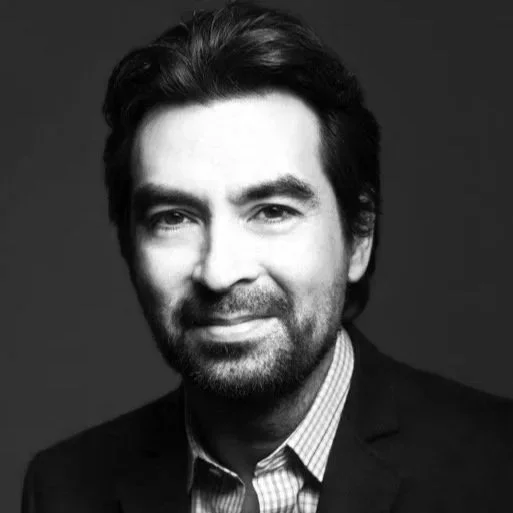}}]{Alexander Gray}
received bachelor's degrees in applied mathematics
and computer science from the University of California, Berkeley,
CA, USA, and the Ph.D. degree in computer science from Carnegie
Mellon University, Pittsburgh, PA, USA, in 2003. He is currently
President and Scientific Director of the Centaur AI Institute,
Lincoln, CA, USA. His research interests include neurosymbolic
artificial intelligence, machine learning, statistical computing,
and scalable algorithms for data analysis.
\end{IEEEbiography}

\Needspace{1.45in}
\begin{IEEEbiography}[{\biographyphoto{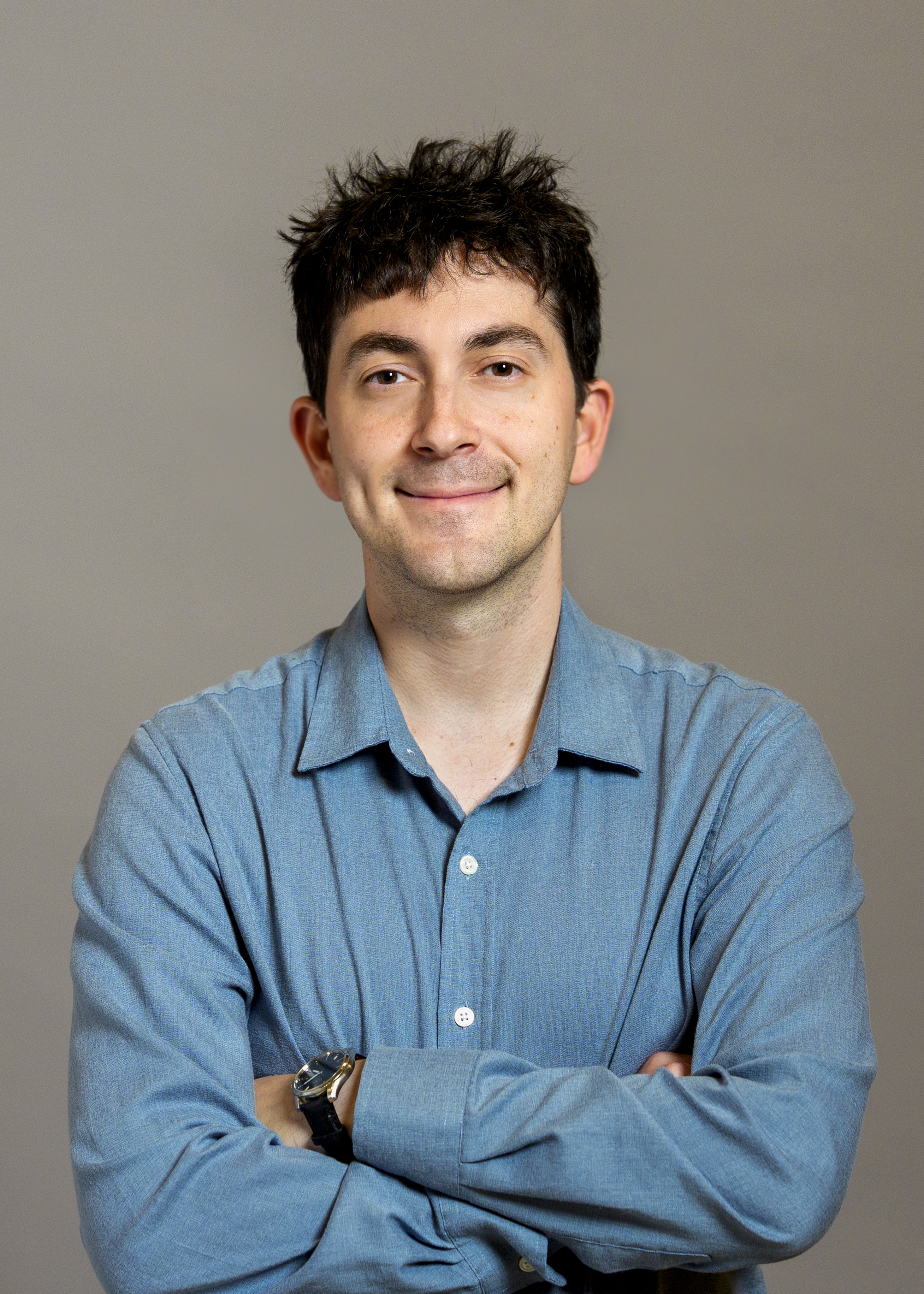}}]{Tom Silver}
received the B.A. degree in computer science and
mathematics from Harvard University, Cambridge, MA, USA, in 2016,
and the Ph.D. degree in electrical engineering and computer science
from the Massachusetts Institute of Technology, Cambridge, MA, USA,
in 2024. He completed postdoctoral research at Cornell University.
He is currently an Assistant Professor of Electrical and Computer
Engineering at Princeton University, Princeton, NJ, USA, where he
directs the Princeton Robot Planning and Learning Lab. His research
focuses on robot planning and machine learning.
\end{IEEEbiography}

\Needspace{1.45in}
\begin{IEEEbiography}[{\biographyphoto{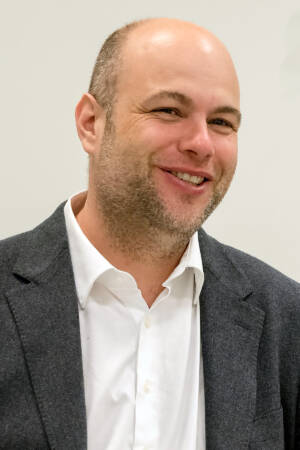}}]{Sebastian Scherer}
received the B.S. degree in computer science and
the M.S. and Ph.D. degrees in robotics from Carnegie Mellon
University, Pittsburgh, PA, USA, in 2004, 2007, and 2010,
respectively. He is currently a Research Professor in the Robotics
Institute at Carnegie Mellon University. His research focuses on
resilient robotic autonomy in challenging environments, including
perception, planning, navigation, aerial robotics, and off road
autonomous systems.
\end{IEEEbiography}

\Needspace{1.45in}
\begin{IEEEbiography}[{\biographyphoto{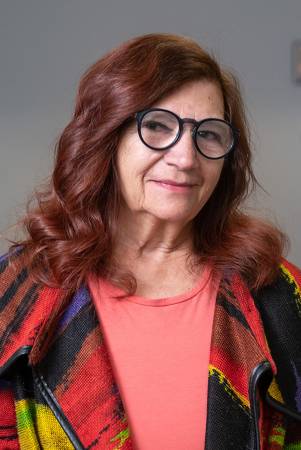}}]{Katia Sycara}
received the B.S. degree in applied mathematics from
Brown University, Providence, RI, USA, the M.S. degree in electrical
engineering from the University of Wisconsin, and the Ph.D. degree
in computer science from Georgia Institute of Technology, Atlanta,
GA, USA. She is the Edward Fredkin Research Professor and Associate
Director of Faculty in the Robotics Institute at Carnegie Mellon
University, Pittsburgh, PA, USA, where she directs the Advanced
Agent Robotics Technology Lab. Her research includes multiagent
systems, multirobot coordination, human robot interaction, and
artificial intelligence. She is a Fellow of IEEE and AAAI.
\end{IEEEbiography}

\Needspace{1.45in}
\begin{IEEEbiography}[{\biographyphoto{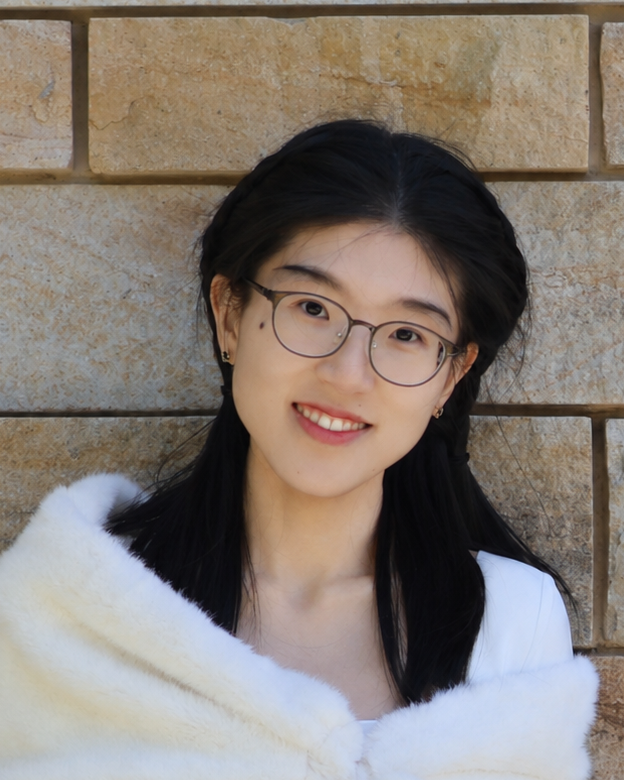}}]{Yaqi Xie}
 received the B.Eng. degree in engineering systems and
design from the Singapore University of Technology and Design,
Singapore, in 2017, and the Ph.D. degree in computer science from
the National University of Singapore, Singapore, in 2023. 
She is currently a Postdoctoral Fellow with the Robotics Institute, Carnegie Mellon University, Pittsburgh, PA, USA. In November 2026, she will join the Department of Electrical and Computer Engineering, University of Illinois Urbana-Champaign, Urbana, IL, USA, as an Assistant Professor. 
Her research interests include robot learning, trustworthy intelligent systems, and foundation models for embodied agents.

\end{IEEEbiography}
\par
\vfill
\endgroup